%% file: main.tex
\documentclass{article} 
\usepackage{iclr2025_conference,times}

\usepackage{wrapfig}
\usepackage{lipsum} 
\usepackage{wrapfig}

\input{math_commands.tex}

\usepackage{booktabs}      
\usepackage{multirow}      
\usepackage{amssymb}       
\usepackage{graphicx, subfigure}     

\usepackage{colortbl}
\usepackage{xcolor}

\usepackage{hyperref}
\usepackage{url}
\usepackage{multicol}
\usepackage{aliascnt}
\usepackage{adjustbox}
\usepackage{siunitx} 
\usepackage{booktabs}
\usepackage{multirow} 
\usepackage{enumitem}
\usepackage{booktabs} 
\usepackage{amssymb}  
\usepackage{amsmath}  
\usepackage{graphicx} 
\usepackage{longtable} 
\usepackage{graphicx}
\usepackage{subcaption}
\usepackage{calc}
\usepackage[most]{tcolorbox}
\usepackage{listings}
\usepackage{latexsym}
\usepackage{dsfont}
\usepackage{mathtools}
\usepackage{makecell}

\usepackage[ruled]{algorithm2e}

\definecolor{uclablue}{rgb}{0.15, 0.45, 0.68}
\hypersetup{
    breaklinks,
    citecolor=uclablue,
    colorlinks=true,
}

\definecolor{lightgreen}{RGB}{0,150,0}
\definecolor{myred}{RGB}{200,0,0} 

\newcommand{\cimp}[2]{%
  \ensuremath{%
    #1%
    \mathrlap{_{\scriptscriptstyle\textcolor{lightgreen}{#2}}}%
  }%
}

\newcommand{\cdec}[2]{%
  \ensuremath{%
    #1%
    \mathrlap{_{\scriptscriptstyle\textcolor{myred}{#2}}}%
  }%
}

\usepackage{tabularx}
\robustify\bfseries
\usepackage{subcaption}
\usepackage{cleveref}

\tcbuselibrary{listings,skins,breakable}

\newtcolorbox{prompt}[1][]{enhanced,
  breakable,
  colback=white,              
  colframe=black,             
  coltitle=black,             
  colbacktitle=gray!20,       
  fonttitle=\bfseries,
  title=Prompt,
  #1}

\newtcolorbox{remark}[1][]{enhanced,
  breakable,
  colback=violet!6,           
  colframe=violet!50!black,   
  coltitle=black,             
  colbacktitle=violet!18,     
  fonttitle=\bfseries,
  title=Remark,
  #1}

\definecolor{linkColor}{rgb}{0.2,0.4,0.6}
\definecolor{myblue}{HTML}{0379AC}
\definecolor{myred}{HTML}{A50E50}
\definecolor{myorange}{RGB}{238, 133, 74}
\definecolor{latentcolor}{named}{cyan}
\definecolor{normalcolor}{RGB}{0, 0, 0}
\usepackage{marvosym}

\title{Learning beyond Teacher: Generalized On-Policy Distillation\\ with Reward Extrapolation}

\author{
Wenkai Yang$^{1,}$\thanks{Work done during an internship at Tencent.} \ , Weijie Liu$^{2}$, Ruobing Xie$^{2}$, Kai Yang$^{2}$, \\ Saiyong Yang$^2$, Yankai Lin$^{1,}$\thanks{Corresponding author.}\\ 
\textbf{$^1$Gaoling School of Artificial Intelligence, Renmin University of China} \\ \textbf{$^2$LLM Department, Tencent}\\
\Letter~\{wenkaiyang,yankailin\}@ruc.edu.cn\\
}

\begin{document}
\maketitle
\let\oldthefootnote\thefootnote

\let\thefootnote\oldthefootnote

\begin{abstract}
On-policy distillation (OPD), which aligns the student with the teacher's logit distribution on student-generated trajectories, has demonstrated strong empirical gains in improving student performance and often outperforms off-policy distillation and reinforcement learning (RL) paradigms. In this work, we first theoretically show that OPD is a special case of dense KL-constrained RL where the reward function and the KL regularization are always weighted equally and the reference model can by any model. Then, we propose the \textbf{Generalized On-Policy Distillation} (\textbf{G-OPD}) framework, which extends the standard OPD objective by introducing a flexible reference model and a reward scaling factor that controls the relative weight of the reward term against the KL regularization. Through comprehensive experiments on math reasoning and code generation tasks, we derive two novel insights: (1) Setting the reward scaling factor to be greater than 1 (i.e., \emph{reward extrapolation}), which we term \textbf{ExOPD}, consistently improves over standard OPD across a range of teacher-student size pairings. In particular, in the setting where we merge the knowledge from different domain experts, obtained by applying domain-specific RL to the same student model, back into the original student, \textbf{ExOPD enables the student to even surpass the teacher's performance boundary and outperform the domain teachers}. (2) Building on ExOPD, we further find that in the strong-to-weak distillation setting (i.e., distilling a smaller student from a larger teacher), performing \emph{reward correction} by choosing the reference model as the teacher's base model before RL yields a more accurate reward signal and further improves distillation performance. However, this choice assumes access to the teacher's pre-RL variant and incurs more computational overhead. We hope our work offers new insights for future research on OPD.\footnote{Code is available at \url{https://github.com/RUCBM/G-OPD}.}~\looseness=-1
\end{abstract}

\begin{figure*}[h]
  \centering
  \vskip -0.1in
  \subfigure[Multi-teacher distillation results, student model is Qwen3-4B-Non-Thinking, teachers are domain-specific RL variants \label{fig: multi-teacher distillation results}]{\includegraphics[width=0.48\textwidth]{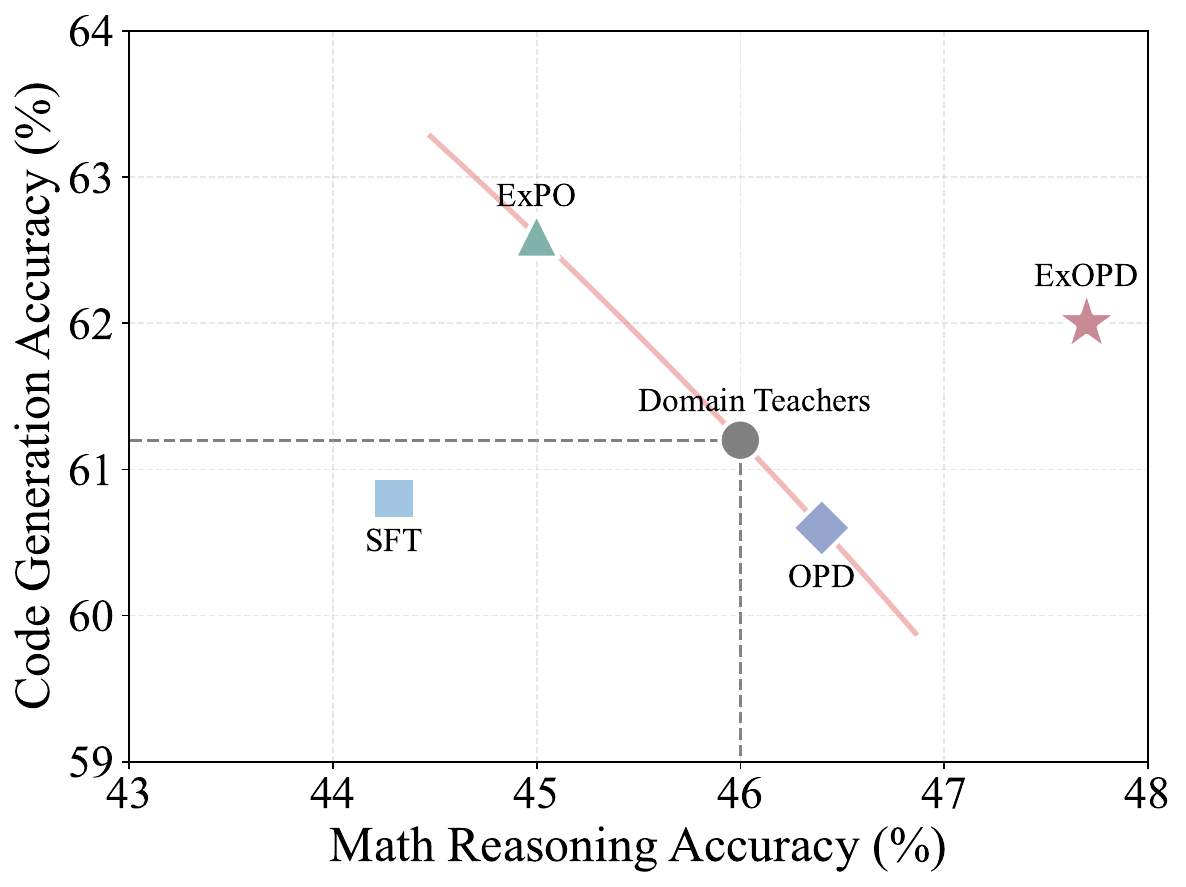}
  }
  \hfill
  \subfigure[Strong-to-weak distillation results, teacher model is Qwen3-30B-A3B-Instruct-2507 \label{fig: strong-to-weak distillation results}]{\includegraphics[width=0.48\textwidth]{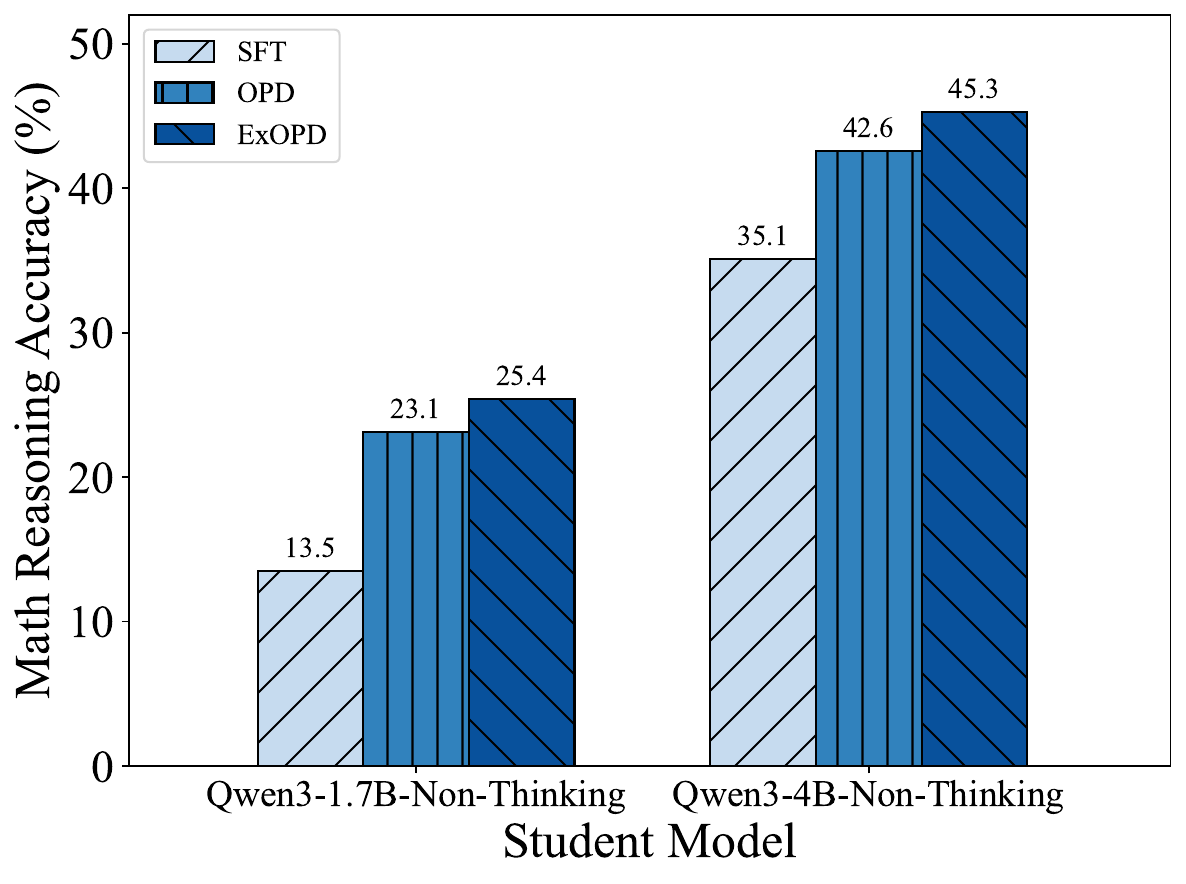}
  }
  \caption{The empirical effectiveness of our method \textbf{ExOPD} compared with off-policy distillation (SFT), standard OPD, and the weight-extrapolation method ExPO~\citep{expo} in multi-teacher and strong-to-weak distillation settings (results averaged over 4 math reasoning and 3 code generation benchmarks). (a) When merging multiple domain experts—obtained by applying domain-specific RL to the same base model—back into the original base model, ExOPD is the only method that \textbf{yields a unified student that consistently outperforms all domain teachers}. (b) ExOPD also yields significant improvements over standard OPD when distilling a smaller student from a larger teacher. Moreover, applying reward correction in ExOPD can further boost distillation performance (Figure~\ref{fig: effect of reward correction}).
  }
  \label{fig: overview}
  \vskip -0.1in
\end{figure*}

\section{Introduction}
Recently, on-policy distillation (OPD)~\citep{opd-iclr,qwen3,opd-tml} has emerged as an effective post-training paradigm for improving capabilities of Large Language Models (LLMs). Unlike prior off-policy distillation methods~\citep{alpaca,openthoughts} that train the student on teacher-generated trajectories, OPD allows the student to learn from the teacher's supervision (i.e., predicted logits) on \emph{student}-generated tokens. Previous studies have shown that OPD can not only serve as a promising multi-task post-training paradigm to (near-)losslessly merge the capabilities acquired by different RL variants across domains back into the original base model~\citep{mimo}, but also be effective and efficient in distilling the capabilities of a larger teacher into a smaller student~\citep{minillm,qwen3}. 

Despite its empirical effectiveness, a mechanistic understanding of OPD remains limited in the field, leaving its full potential under-explored. In this work, we bridge this gap by establishing a theoretical connection between OPD and dense reinforcement learning (RL), and by extending standard OPD into a generalized formulation.

First, we make derivations to show that OPD is essentially a special case of the standard dense RL with Kullback–Leibler (KL) constraint, where the token-level reward function is always weighted equally with the KL regularization and the reference model can be chosen arbitrarily. Building on this insight, we generalize the OPD objective to a more universal formulation by further introducing a \textit{reward scaling factor} that controls the relative weight of the reward term against the KL regularization, in addition to the \textit{flexible reference model}. We refer to this generalized formulation as the \textbf{Generalized On-Policy Distillation} (\textbf{G-OPD}) framework.

Based on the G-OPD framework, we theoretically analyze how the reward scaling factor and the choice of reference model affect distillation effectiveness across different settings, supported by comprehensive experiments in both math reasoning and code generation domains. 
In the first setting, the teacher is obtained by applying domain-specific RL to the student, and the reference model is naturally fixed to the student's initial state. We show that (1) when the reward scaling factor lies in $(0,1)$ (i.e., \textit{reward interpolation}), the distilled student exhibits behaviors (e.g., performance and response length) that fall between the reference and teacher models; (2) when the reward scaling factor is greater than $1$ (i.e., \textit{reward extrapolation}), the student can learn beyond the teacher's capability boundary and outperform teacher in domain tasks. We refer to the reward extrapolation variant as \textbf{ExOPD}. We further show that ExOPD extends well to the multi-teacher distillation setting, \textbf{enabling a unified student to surpass all domain teachers}. Second, we study the strong-to-weak distillation setting, where a smaller student is distilled from a larger teacher. In this setting, we demonstrate that replacing the reference model from the student's initial policy to the teacher's pre-RL variant (i.e., \emph{reward correction}) in ExOPD yields a more accurate reward signal and further improves distillation performance. However, the limitations of this practice are that it assumes access to an additional model (the teacher's pre-RL variant) and incurs more computational cost on computing the log-probabilities of the larger reference model. Despite these limitations, ExOPD and ExOPD with reward correction significantly outperform standard OPD in the strong-to-weak distillation setting.

\section{Related Work}

\textbf{Off-Policy Distillation.} Knowledge distillation (KD)~\citep{kd} is a widely used technique for transferring knowledge from a domain expert (teacher) to a student model. 
Most prior studies focus on \emph{off-policy} distillation, where the student is trained on trajectories generated by the teacher, either by aligning the student's logits distribution with the teacher's via a Kullback–Leibler (KL) divergence loss on token logits~\citep{distilbert,seq-kd,leaf}, or by directly performing supervised fine-tuning (SFT) with a cross-entropy loss on the teacher-generated tokens~\citep{alpaca,lima,openthoughts}. This practice has been shown to effectively improve the student model across a broad range of capabilities~\citep{ultrachat,tops,limo} in the LLM era. 


\textbf{On-Policy Distillation.} 
By sampling trajectories from the student and aligning the student with the teacher's logit distribution on each token of these student-generated trajectories, on-policy distillation (OPD)~\citep{opd-iclr,minillm} realizes \emph{dense on-policy} learning. 
Empirically, OPD has been shown to achieve faster and more effective distillation than off-policy distillation~\citep{qwen3,opd-tml}. Recent OPD studies have explored distillation across different model families~\citep{gold}, developed black-box on-policy distillation methods that do not require access to the teacher's logits~\citep{black-box-opd}, and investigated the self-distillation paradigm that leverage the LLM's in-context capabilities to distill textual context information into its parameters~\citep{rule-distillation,SDPO,self-distillation-continual-learning,self-distilled-reasoning,pi-privileged}.~\looseness=-1

\section{Methodology}
\subsection{Preliminaries}
In this section, we start with a brief review of relevant preliminaries.

\textbf{Off-Policy Distillation.} 
Let $D$ denote the input distribution, and let $\pi_{\boldsymbol{\theta}}$ and $\pi^{*}$ denote the student and teacher policies, respectively. The general form of Knowledge Distillation (KD)~\citep{kd} can be written as
\begin{equation}
\label{eq:kd}
\mathcal{J}_{\text{KD}}(\boldsymbol{\theta}) =  \min_{\boldsymbol{\theta}}\;
\mathbb{E}_{\boldsymbol{x} \sim D , \boldsymbol{y} \sim \pi^{*}(\cdot | \boldsymbol{x}) }\Big[
\mathcal{D}_{\mathrm{KL}} \!\big(
\pi^{*}(\boldsymbol{y} | \boldsymbol{x})
\,\big\|\,
\pi_{\boldsymbol{\theta}}( \boldsymbol{y} | \boldsymbol{x})
\big)
\Big],
\end{equation}
where $\mathcal{D}_{\mathrm{KL}}$ denotes the Kullback--Leibler (KL) divergence loss. 
In the era of LLMs, obtaining the teacher's full output distribution (e.g., logits) is often expensive or even infeasible. As a result, KD is commonly implemented as supervised fine-tuning (SFT) of the student on trajectories generated by the teacher. Though effective, \textbf{the major drawback of this paradigm is its \textit{off-policy} nature}: the student is trained to imitate the teacher's behavior, rather than to learn from reward signals induced by its own actions. As a result, it may fail to adapt and generalize from its own experience at test time, when faced with similar problems.

\textbf{On-Policy RL.} We use $\pi_{\boldsymbol{\theta}}$ to denote the policy model to be optimized. The RL objective can be formulated as
\begin{equation}
\label{eq:rl_objective_general}
\mathcal{J}_{\text{RL}}(\boldsymbol{\theta}) =  \max_{\boldsymbol{\theta}}\;
\mathbb{E}_{\boldsymbol{x} \sim D, \boldsymbol{y} \sim \pi_{\boldsymbol{\theta}}(\cdot|x)}\Big[
r(\boldsymbol{x},\boldsymbol{y}) - \beta \mathcal{D}_{\mathrm{KL}}(\pi_{\boldsymbol{\theta}}\,\|\, \pi_{\mathrm{ref}})
\Big].
\end{equation}
In the above formulation, the trajectories $y$ are sampled from the current policy model, making the training remain \textit{on-policy}. $r(\boldsymbol{x},\boldsymbol{y})$ is the reward function that measures the quality of a response sequence $\boldsymbol{y}=(y_1,\cdots, y_{T})$ to a query $\boldsymbol{x}$. Depending on the setting, it can be either (i) a parameterized neural reward model trained on the specific preference data for open-domain alignment~\citep{internlm2,rlhf-workflow,skywork-rm-v2}, or (ii) a rule-based, deterministic outcome verifier commonly used in verifiable LLM reasoning tasks~\citep{r1,orz,code-r1,deepcritic}. $\mathcal{D}_{\mathrm{KL}}(\pi_{\boldsymbol{\theta}} \,\|\, \pi_{\mathrm{ref}})$ prevents the policy model $\pi_{\boldsymbol{\theta}}$ from drifting too far from a reference model $\pi_{\mathrm{ref}}$, and the coefficient $\beta$ controls the strength of this constraint. To solve Eq.~(\ref{eq:rl_objective_general}), a common approach is to apply policy gradient~\citep{rl_introduction}, updating the policy parameters using an estimated gradient of the form
\begin{equation}
\label{eq:policy_gradient_general}
\nabla_{\boldsymbol{\theta}}\mathcal{J}_{\text{RL}}(\boldsymbol{\theta})
=
\mathbb{E}_{\boldsymbol{x}\sim D,\;\boldsymbol{y}\sim \pi_{\boldsymbol{\theta}}(\cdot | \boldsymbol{x})}
\Big[ \sum_{t=1}^{T}
A_{t} \nabla_{\boldsymbol{\theta}} \log \pi_{\boldsymbol{\theta}} (y_{t} | \boldsymbol{x},\boldsymbol{y}_{<t})
\Big],
\end{equation}
where $A_t$ is the relative advantage of token $y_t$ over a baseline value. In practice, \textbf{the reward signal in RL is often sparse}: the policy model only receives a reward at the final token after the response is completed, which may make optimization inefficient and ineffective~\citep{prime}.

\textbf{On-Policy Distillation.} On-Policy Distillation (OPD)~\citep{opd-iclr,minillm,opd-tml} inherits the \textit{on-policy} nature of policy training and the advantage of \textit{dense credit assignment}, making it an efficient post-training paradigm~\citep{qwen3,mimo}. The main idea of OPD is to let the student generate its own trajectories, and then minimize the reverse KL divergence between the student and the teacher $\pi^{*}$ on those student-generated trajectories:
\begin{equation}
\label{eq:opd}
\mathcal{J}_{\text{OPD}}(\boldsymbol{\theta}) =  \min_{\boldsymbol{\theta}}\;
\mathbb{E}_{\boldsymbol{x} \sim D, \boldsymbol{y} \sim \pi_{\boldsymbol{\theta}} (\cdot | \boldsymbol{x}) }\Big[
\mathcal{D}_{\mathrm{KL}} \!\Big(
\pi_{\boldsymbol{\theta}}(\boldsymbol{y} | \boldsymbol{x})
\,\big\|\,
\pi^{*}( \boldsymbol{y} | \boldsymbol{x})
\Big)
\Big].
\end{equation}
Notice that in Eq.~(\ref{eq:opd}), the trajectories $\boldsymbol{y}$ are generated by the policy model itself, resulting in the on-policy training. Also, we can get the gradient of OPD as\footnote{Detailed derivations are in Appendix~\ref{appendix: math derivations}.}
\begin{equation}
\label{eq:opd_gradient_original}
\nabla_{\boldsymbol{\theta}} \mathcal{J}_{\text{OPD}}(\boldsymbol{\theta})=\mathbb{E}_{\boldsymbol{x} \sim D, \boldsymbol{y} \sim \pi_{\boldsymbol{\theta}} (\cdot | \boldsymbol{x}) }\Big[
\sum_{t=1}^{T} \Big( \sum_{t^{'} =t}^{T} \big(\log \pi_{\boldsymbol{\theta}}(y_{t^{'}}| \boldsymbol{x}, \boldsymbol{y}_{<t^{'}} ) - \log \pi^{*}(y_{t^{'}}| \boldsymbol{x}, \boldsymbol{y}_{<t^{'}} ) \big) \Big) \, \nabla_{\boldsymbol{\theta}} \log \pi_{\boldsymbol{\theta}}(y_t| \boldsymbol{x}, \boldsymbol{y}_{<t} ) 
\Big].
\end{equation}
In practice, current studies~\citep{opd-tml,mimo} use a discount factor of 0 (focus on next-token optimization only) and approximate the gradient as
\begin{equation}
\label{eq:opd_gradient}
\nabla_{\boldsymbol{\theta}} \mathcal{J}_{\text{OPD}}(\boldsymbol{\theta}) =  
\mathbb{E}_{\boldsymbol{x} \sim D, \boldsymbol{y} \sim \pi_{\boldsymbol{\theta}} (\cdot | \boldsymbol{x}) }\Big[
\sum_{t=1}^{T} \big(\log \pi_{\boldsymbol{\theta}}(y_t| \boldsymbol{x}, \boldsymbol{y}_{<t} ) - \log \pi^{*}(y_t| \boldsymbol{x}, \boldsymbol{y}_{<t} ) \big) \, \nabla_{\boldsymbol{\theta}} \log \pi_{\boldsymbol{\theta}}(y_t| \boldsymbol{x}, \boldsymbol{y}_{<t} ) 
\Big].
\end{equation}
Comparing Eq.~(\ref{eq:opd_gradient}) with Eq.~(\ref{eq:policy_gradient_general}), we can see that $-\big(\log \pi_{\boldsymbol{\theta}}(y_t| \boldsymbol{x}, \boldsymbol{y}_{<t} ) - \log \pi^{*}(y_t| \boldsymbol{x}, \boldsymbol{y}_{<t} )\big)$ can be regarded as the token-level advantage in OPD, thereby providing dense credit assignment for each token-level action. 

\subsection{Generalized On-Policy Distillation}
In this section, we first start from Eq.~(\ref{eq:opd}) and derive a generalized formulation of OPD. 

First, we re-formulate the OPD objective~\citep{il_rlhf} as
\begin{equation}
\label{eq:op_derivation}
\begin{aligned}
\mathcal{J}_{\text{OPD}}(\boldsymbol{\theta}) &=  \min_{\boldsymbol{\theta}}\;
\mathbb{E}_{\boldsymbol{x} \sim D, \boldsymbol{y} \sim \pi_{\boldsymbol{\theta}} (\cdot | \boldsymbol{x}) }\Big[
\mathcal{D}_{\mathrm{KL}} \!\big(
\pi_{\boldsymbol{\theta}}(\boldsymbol{y} | \boldsymbol{x})
\,\big\|\,
\pi^{*}( \boldsymbol{y} | \boldsymbol{x})
\big)
\Big] \\ & = \min_{\boldsymbol{\theta}}\;
\mathbb{E}_{\boldsymbol{x} \sim D, \boldsymbol{y} \sim \pi_{\boldsymbol{\theta}} (\cdot | \boldsymbol{x}) }\Big[
\log \pi_{\boldsymbol{\theta}}(\boldsymbol{y}| \boldsymbol{x} ) - \log \pi^{*}(\boldsymbol{y}| \boldsymbol{x} )
\Big] \\ & =
\max_{\boldsymbol{\theta}}\;
\mathbb{E}_{\boldsymbol{x} \sim D, \boldsymbol{y} \sim \pi_{\boldsymbol{\theta}} (\cdot | \boldsymbol{x}) }\Big[
\log \pi^{*}(\boldsymbol{y}| \boldsymbol{x} ) - \log \pi_{\boldsymbol{\theta}}(\boldsymbol{y}| \boldsymbol{x} ) \Big]  \\ &=
\max_{\boldsymbol{\theta}}\;
\mathbb{E}_{\boldsymbol{x} \sim D, \boldsymbol{y} \sim \pi_{\boldsymbol{\theta}} (\cdot | \boldsymbol{x}) }\Big[
\big(\log \pi^{*}(\boldsymbol{y}| \boldsymbol{x} ) - \log \pi_{\mathrm{ref}} (\boldsymbol{y}| \boldsymbol{x} ) \big) - \big( \log \pi_{\boldsymbol{\theta}}(\boldsymbol{y}| \boldsymbol{x} ) - \log \pi_{\mathrm{ref}} (\boldsymbol{y}| \boldsymbol{x} ) \big)
\Big] \\ & =
\max_{\boldsymbol{\theta}}\;
\mathbb{E}_{\boldsymbol{x} \sim D, \boldsymbol{y} \sim \pi_{\boldsymbol{\theta}} (\cdot | \boldsymbol{x}) }\Big[
\log  \frac{\pi^{*}(\boldsymbol{y}| \boldsymbol{x} )}{\pi_{\mathrm{ref}} (\boldsymbol{y}| \boldsymbol{x} )}  - \mathcal{D}_{\mathrm{KL}} \!\big(
\pi_{\boldsymbol{\theta}}(\boldsymbol{y} | \boldsymbol{x})
\,\big\|\,
\pi_{\mathrm{ref}}( \boldsymbol{y} | \boldsymbol{x})
\big)
\Big].
\end{aligned}
\end{equation}
Therefore, we have the following remark:
\begin{remark}
    By introducing a third reference model $\pi_{\mathrm{ref}}$, the OPD objective in Eq.~(\ref{eq:opd}) becomes equivalent to a specific KL-constrained RL objective in Eq.~(\ref{eq:rl_objective_general}), where the reward function $r(\boldsymbol{x},\boldsymbol{y})=\log \frac{\pi^{*}(\boldsymbol{y} | \boldsymbol{x})}{\pi_{\mathrm{ref}}(\boldsymbol{y} | \boldsymbol{x})}$, the KL regularization is applied between the policy model $\pi_{\boldsymbol{\theta}}$ and the reference model $\pi_{\mathrm{ref}}$, and the reward and KL terms are weighted equally (i.e., $\beta=1$ in Eq.~(\ref{eq:rl_objective_general})).
\end{remark}
From the above remark, we establish the connection between OPD and RL. However, we emphasize that OPD differs from standard RL in the following key respects:

\textbf{(1) Dense rewards.} As discussed above, in standard RL the model typically receives an effective reward only at the final token, while the rewards for all other tokens are zero:
\begin{equation}
\label{eq:rewards_in_rl}
r_{t}^{RL}= \begin{cases}
0 & t=1,\cdots,T-1 , \\
\text{Outcome Reward} & t=T .
\end{cases}
\end{equation}
However, in OPD, each token-level action receives an effective reward 
\begin{equation}
\label{eq:rewards_in_opd}
r_{t}^{OPD}= \log \frac{\pi^{*}(y_t | \boldsymbol{x}, \boldsymbol{y}_{<t})}{\pi_{\mathrm{ref}}(y_t | \boldsymbol{x}, \boldsymbol{y}_{<t})}, \quad t=1,\cdots, T.
\end{equation}
This token-level reward takes essentially the same form as the \textit{implicit reward} defined in \citet{dpo}. Implicit reward is initially derived from the closed-form solution of Eq.~(\ref{eq:rl_objective_general}), which can be written as
\begin{equation}
\label{eq:rl_solution}
r(\boldsymbol{x},\boldsymbol{y})=\beta \log \frac{\pi_{\boldsymbol{\theta}}(\boldsymbol{y}|\boldsymbol{x})}{ \pi_{\mathrm{ref}}(\boldsymbol{y}|\boldsymbol{x})} + \beta \log Z(\boldsymbol{x}),\text{ where } Z(\boldsymbol{x})=\sum_{\boldsymbol{y}} \pi_{\text{ref}}(\boldsymbol{y}|\boldsymbol{x}) \exp (\frac{1}{\beta}r(\boldsymbol{x},\boldsymbol{y})).
\end{equation}
As we can see, since $\log Z(\boldsymbol{x})$ is a constant depending only on $\boldsymbol{x}$,  $\log \frac{\pi_{\boldsymbol{\theta}}(\boldsymbol{y}|\boldsymbol{x})}{ \pi_{\mathrm{ref}}(\boldsymbol{y}|\boldsymbol{x})}$ can be regarded as a well-defined proxy of the true reasoning reward, and this idea is adopted in previous studies~\citep{implicitprm,prime,laser,agentic_implicit_reward} to provide dense supervision for RL. However, in OPD, the implicit reward
$\log \frac{\pi^{*}(\boldsymbol{y} | \boldsymbol{x} )}{\pi_{\mathrm{ref}}(\boldsymbol{y} | \boldsymbol{x})}$
does not require $\pi^{*}$ to be obtained by applying RL starting from $\pi_{\mathrm{ref}}$. In fact, $\pi^{*}$ and $\pi_{\mathrm{ref}}$ can even be models of different sizes. Nevertheless, \textbf{this reward function still captures the log-probability shift from the reference ($\pi_{\mathrm{ref}}$) distribution to the expert ($\pi^{*}$) distribution}, and thus provides a meaningful training signal.

\textbf{(2) Fixed weighting between the reward function and the KL regularization.} As revealed in the remark, in OPD, the reward term and the KL regularization are always weighted equally. In what follows, we present and discuss our generalized OPD formulation by introducing a reward scaling factor that allows us to adjust the relative weight of the reward term against the KL regularization.

\textbf{(3) Flexible choice of the reference model.} In RL (i.e., Eq.~(\ref{eq:rl_objective_general})), the reference model is typically initialized as the policy model's starting checkpoint. However, we note that in OPD (i.e., Eq.~(\ref{eq:generalized_opd})), the introduced reference model can be any model, since this choice does not affect the final simplification of the objective back to its original form in Eq.~(\ref{eq:opd}). In what follows, we discuss how different choices of $\pi_{\mathrm{ref}}$ affect our proposed generalized OPD framework. By default, the reference model is selected as the student's initial policy.

From the above discussion, we can see that OPD offers two key advantages over RL—dense reward signals and a flexible choice of reference model—yet it fixes the relative weighting between the reward function and the KL regularization to $1:1$. This motivates us to follow Eq.~(\ref{eq:rl_objective_general}) and generalize the original OPD objective in Eq.~(\ref{eq:opd}) into a general dense RL objective with a flexible KL constraint, by introducing both a third reference model and an additional \textbf{reward scaling factor $\lambda$}:
\begin{equation}
\label{eq:generalized_opd}
\begin{aligned}
\mathcal{J}_{\text{G-OPD}}(\boldsymbol{\theta}) & =
\max_{\boldsymbol{\theta}}\;
\mathbb{E}_{\boldsymbol{x} \sim D, \boldsymbol{y} \sim \pi_{\boldsymbol{\theta}} (\cdot | \boldsymbol{x}) }\Big[ \textcolor{red}{\lambda}
\log  \frac{\pi^{*}(\boldsymbol{y}| \boldsymbol{x} )}{\textcolor{blue}{\pi_{\mathrm{ref}}} (\boldsymbol{y}| \boldsymbol{x} )}  - \mathcal{D}_{\mathrm{KL}} \!\big(
\pi_{\boldsymbol{\theta}}(\boldsymbol{y} | \boldsymbol{x})
\,\big\|\,
\textcolor{blue}{\pi_{\mathrm{ref}}}( \boldsymbol{y} | \boldsymbol{x})
\big)
\Big].
\end{aligned}
\end{equation}
The above Eq.~(\ref{eq:generalized_opd}) presents our \textbf{Generalized On-Policy Distillation} (\textbf{G-OPD}) formulation, where $\lambda$ controls the relative weight of the reward term against the KL regularization in the objective, and is essential $\frac{1}{\beta}$ in Eq.~(\ref{eq:rl_objective_general}). As we can see, compared to RL, G-OPD enables dense credit assignment and a more flexible choice of reference model; compared to OPD, it further allows more general control over the reward weight. In the following, we discuss in detail about the two crucial components, $\lambda$ and $\pi_{\mathrm{ref}}$, in G-OPD.

\paragraph{Reward interpolation and extrapolation in G-OPD.} The optimal solution to G-OPD in Eq.~(\ref{eq:generalized_opd}) satisfies that 
\begin{equation}
\label{eq:g-opd optimal solution}
\begin{aligned}
\log \pi_{\boldsymbol{\theta}} (\boldsymbol{y} | \boldsymbol{x}) & = \lambda \log  \pi^{*} (\boldsymbol{y} | \boldsymbol{x}) + (1-\lambda) \log \pi_{\mathrm{ref}} (\boldsymbol{y} | \boldsymbol{x}) \\ & = \log  \pi^{*} (\boldsymbol{y} | \boldsymbol{x})  + (\lambda - 1) (\log  \pi^{*} (\boldsymbol{y} | \boldsymbol{x})  - \log  \pi_{\mathrm{ref}} (\boldsymbol{y} | \boldsymbol{x}) ).
\end{aligned}
\end{equation}
This reveals that, (1) when $0<\lambda < 1$, G-OPD encourages the student model's log-probability distribution to match a linear interpolation between that of the teacher and reference models. This can also be interpreted as replacing the reward $r$ in Eq.~(\ref{eq:op_derivation}) with $\lambda \cdot r + (1-\lambda)\cdot 0$. Therefore, we refer to this case as \emph{\textbf{reward interpolation}}. We conjecture that, under this setting, the student trained with G-OPD may exhibit behavior (e.g., performance, response length, etc.) that lies between the reference model and the standard OPD with $\lambda=1$. (2) When $\lambda>1$, G-OPD encourages the student's log-probability distribution to go beyond matching the teacher's log-probabilities by additionally fitting an extra shift term $(\lambda - 1)(\log \pi^{*} - \log \pi_{\mathrm{ref}})$. From the perspective of rewards, G-OPD with $\lambda > 1$ performs an extrapolation of the reward function's weight in Eq.~(\ref{eq:op_derivation}); thus, we refer to this regime as \emph{\textbf{reward extrapolation}}. We wonder whether reward extrapolation can outperform standard OPD, and in a special case, when the teachers are domain experts obtained by applying RL to the same student~\citep{mimo} in different domains, \textbf{can reward extrapolation in G-OPD distill a unified student that surpasses all the domain teachers?}

\paragraph{Reward correction in strong-to-weak distillation.} When the reward scaling factor $\lambda \neq 1$, different choices of the reference model $\pi_{\mathrm{ref}}$ in Eq.~(\ref{eq:generalized_opd}) lead to different objectives. Based on distillation settings, in the following, we discuss the choices of $\pi_{\mathrm{ref}}$ in two cases: (1) One application of G-OPD is to merge the capabilities of several experts, each obtained by applying domain-specific RL starting from the same base model, back into the original base model~\citep{mimo}. In this setting, $\pi_{\mathrm{ref}}$ is naturally chosen as the original base model, and the reward function in G-OPD is exactly the implicit reward defined in Eq.~(\ref{eq:rl_solution}). (2) Another distillation setting is \emph{strong-to-weak distillation}~\citep{qwen3}, i.e., distilling a large teacher into a smaller student. In this case, $\pi_{\mathrm{ref}}$ admits two choices: (i) the student’s base model, $\pi_{\mathrm{base}}^{\mathrm{student}}$, which corresponds to the default setting where we only have access to $\pi^{*}$ and $\pi_{\mathrm{base}}^{\mathrm{student}}$; or (ii) the teacher expert’s pre-RL base model, $\pi_{\mathrm{base}}^{\mathrm{teacher}}$ (i.e., the teacher before post-training), assuming it is available. To compare these two choices, we first rewrite the G-OPD objective into an equivalent form:
\begin{equation}
\label{eq:generalized_opd_2}
\begin{aligned}
\mathcal{J}_{\text{G-OPD}}(\boldsymbol{\theta}) & =
\max_{\boldsymbol{\theta}}\;
\mathbb{E}_{\boldsymbol{x} \sim D, \boldsymbol{y} \sim \pi_{\boldsymbol{\theta}} (\cdot | \boldsymbol{x}) }\Big[ \lambda
\log  \frac{\pi^{*}(\boldsymbol{y}| \boldsymbol{x} )}{\pi_{\mathrm{ref}} (\boldsymbol{y}| \boldsymbol{x} )}  - \mathcal{D}_{\mathrm{KL}} \!\big(
\pi_{\boldsymbol{\theta}}(\boldsymbol{y} | \boldsymbol{x})
\,\big\|\,
\pi_{\mathrm{ref}}( \boldsymbol{y} | \boldsymbol{x})
\big)
\Big] \\ & = 
\max_{\boldsymbol{\theta}}\;
\mathbb{E}_{\boldsymbol{x} \sim D, \boldsymbol{y} \sim \pi_{\boldsymbol{\theta}} (\cdot | \boldsymbol{x}) }\Big[ (\lambda - 1)
\log  \frac{\pi^{*}(\boldsymbol{y}| \boldsymbol{x} )}{\pi_{\mathrm{ref}} (\boldsymbol{y}| \boldsymbol{x} )}  - \mathcal{D}_{\mathrm{KL}} \!\big(
\pi_{\boldsymbol{\theta}}(\boldsymbol{y} | \boldsymbol{x})
\,\big\|\,
\pi^{*}( \boldsymbol{y} | \boldsymbol{x})
\big)
\Big].
\end{aligned}
\end{equation}
Now, under the same KL regularization strength, we can see that choosing $\pi_{\mathrm{ref}}=\pi_{\mathrm{base}}^{\mathrm{teacher}}$ is more reasonable. The reason is that the reward $\log \frac{\pi^{*}}{\pi_{\mathrm{base}}^{\mathrm{teacher}}}$ corresponds to the implicit reward induced by the teacher’s RL post-training, and is thus well-defined according to Eq.~(\ref{eq:rl_solution}). In contrast, $\log \frac{\pi^{*}}{\pi_{\mathrm{base}}^{\mathrm{student}}}$ can be noisier, since there exists fundamental gap between the internal knowledge and capacity of teacher and student base models. Therefore, in the strong-to-weak distillation setting, we think that applying a \emph{\textbf{reward correction}} to the default reward $\log \frac{\pi^{*}}{\pi_{\mathrm{base}}^{\mathrm{student}}}$—by adding $\log \frac{\pi_{\mathrm{base}}^{\mathrm{student}}}{\pi_{\mathrm{base}}^{\mathrm{teacher}}}$ to obtain $\log \frac{\pi^{*}}{\pi_{\mathrm{base}}^{\mathrm{teacher}}}$—can lead to better distillation performance. The limitations, however, are that this requires access to $\pi_{\mathrm{base}}^{\mathrm{teacher}}$ and incurs additional computation, since computing $\log \pi_{\mathrm{base}}^{\mathrm{teacher}}$ requires more cost than computing $\log \pi_{\mathrm{base}}^{\mathrm{student}}$.

\begin{remark}
\label{remark: G-OPD}
    By introducing a reference model $\textcolor{blue}{\pi_{\mathrm{ref}}}$ and a reward scaling factor $\textcolor{red}{\lambda}$, we formulate the Generalized On-Policy Distillation framework as 
    $$
    \mathcal{J}_{\text{G-OPD}}(\boldsymbol{\theta})  =
\max_{\boldsymbol{\theta}}\;
\mathbb{E}_{\boldsymbol{x} \sim D, \boldsymbol{y} \sim \pi_{\boldsymbol{\theta}} (\cdot | \boldsymbol{x}) }\Big[ \textcolor{red}{\lambda}
\log  \frac{\pi^{*}(\boldsymbol{y}| \boldsymbol{x} )}{\textcolor{blue}{\pi_{\mathrm{ref}}} (\boldsymbol{y}| \boldsymbol{x} )}  - \mathcal{D}_{\mathrm{KL}} \!\big(
\pi_{\boldsymbol{\theta}}(\boldsymbol{y} | \boldsymbol{x})
\,\big\|\,
\textcolor{blue}{\pi_{\mathrm{ref}}}( \boldsymbol{y} | \boldsymbol{x})
\big)
\Big].
    $$
We have two observations:

(1) For $0<\lambda<1$, G-OPD yields a student whose behavior lies between that of the reference model and that of the student trained by standard OPD (i.e., $\lambda=1$). In contrast, $\lambda>1$ can potentially deliver larger gains than standard OPD, and may even produce a student that outperforms the teacher in certain cases. Note that setting $\lambda \neq 1$ incurs additional computational cost on computing $\log \pi_{\mathrm{ref}}$. 

(2) In strong-to-weak distillation, choosing the teacher’s pre-RL base model $\pi_{\mathrm{base}}^{\mathrm{teacher}}$ as the reference may yield better distillation performance. However, this comes with two limitations: it requires access to an additional model $\pi_{\mathrm{base}}^{\mathrm{teacher}}$, and it increases computational cost because computing $\log \pi_{\mathrm{base}}^{\mathrm{teacher}}$ requires more computational cost than computing $\log \pi_{\mathrm{base}}^{\mathrm{student}}$.
\end{remark}

Finally, the approximated gradient of G-OPD can be written as
\begin{equation}
\begin{aligned}
\nabla_{\boldsymbol{\theta}} \mathcal{J}_{\text{G-OPD}}(\boldsymbol{\theta}) = 
\mathbb{E}_{\boldsymbol{x} \sim D, \boldsymbol{y} \sim \pi_{\boldsymbol{\theta}} (\cdot | \boldsymbol{x}) }\Big[
\sum_{t=1}^{T} A_{t}^{\text{G-OPD}} \, \nabla_{\boldsymbol{\theta}} \log \pi_{\boldsymbol{\theta}}(y_t| \boldsymbol{x}, \boldsymbol{y}_{<t} ) 
\Big],
\end{aligned}
\end{equation}
where $A_{t}^{\text{G-OPD}} = \big(\log \pi_{\boldsymbol{\theta}}(y_t| \boldsymbol{x}, \boldsymbol{y}_{<t} ) - \log \pi^{*}(y_t| \boldsymbol{x}, \boldsymbol{y}_{<t} ) \big) + (\lambda - 1)\big(\log \pi_{\text{ref}}(y_t| \boldsymbol{x}, \boldsymbol{y}_{<t} ) - \log \pi^{*}(y_t| \boldsymbol{x}, \boldsymbol{y}_{<t} ) \big)$.

\section{Experiments and Analysis}
In this section, we conduct a series of extensive experiments on math reasoning and code generation tasks to analyze the properties of the proposed G-OPD framework and assess the effectiveness of ExOPD. We begin with preliminary experiments on same-size teacher-student pairs in Section~\ref{subsubsec: single-teacher distillation}, where we investigate the impact of the reward scaling factor within G-OPD. We then explore the effectiveness of ExOPD in the multi-teacher distillation setting in Section~\ref{subsubsec: multi-teacher distillation}. Finally, we present experimental results in the strong-to-weak distillation setting in Section~\ref{subsec: strong-to-weak distillation}.

\subsection{Experiments with Same-Sized Student and Teacher}
\label{subsec: same-sized distillation}
Here, we consider the scenario where the domain teachers are reinforced models derived from the student through domain-specific RL. 

\subsubsection{Experimental Settings}
\textbf{Base Model.} We primarily conduct experiments using the Qwen3-4B-Non-Thinking~\citep{qwen3} model. The student model is initialized as Qwen3-4B-Non-Thinking, while the domain teachers are derived by applying RL separately to Qwen3-4B-Non-Thinking on domain-specific data.

\textbf{Training Datasets.} We filter the DeepMath~\citep{deepmath} dataset to select 57K samples with a difficulty level greater than or equal to 6 to form the math RL data, and use Eurus-RL-Code~\citep{prime} as the code RL data, which consists of 25K samples. We then apply RL to the base model on two datasets separately to get domain teachers, Qwen3-4B-Non-Thinking-RL-Math and Qwen3-4B-Non-Thinking-RL-Code. The distillation data is the same as the RL data.

\textbf{Training Settings.} We apply Group Relative Policy Optimization (GRPO)~\citep{deepseekmath} to obtain domain teachers. In RL, a reward of 1.0 is given when the final answer is correct in math reasoning or when all unit tests pass in code generation; otherwise, the reward is 0.0. Detailed training hyper-parameters in GRPO are in Appendix~\ref{appendix: training settings}. After this, we implement G-OPD on the original student model (i.e., Qwen3-4B-Non-Thinking) with different reward scaling factors $\lambda \in \{0.0, 0.25, 0.5, 0.75, 1.0, 1.25, 1.5\}$. Note that $\lambda=0.0$ corresponds to the initial state Qwen3-4B-Non-Thinking, and $\lambda=1.0$ corresponds to standard OPD. The reference model here is fixed naturally as Qwen3-4B-Non-Thinking. Detailed training hyper-parameters in G-OPD are in Appendix~\ref{appendix: training settings}. In both GRPO and G-OPD, we implement token-level rollout correction~\citep{rollout-correction} to mitigate training-inference mismatch. Our experiments are based on \texttt{verl}~\citep{verl} framework.

\textbf{Evaluation.} For the evaluation of math reasoning, we select four competition-level benchmarks: AIME24~\citep{aime2024}, AIME25~\citep{aime2025}, HMMT25 (February)~\citep{hmmt25}, and HMMT25 (November)~\citep{hmmt25}. For the evaluation of code generation, we select three test sets: HumanEval+, MBPP+~\citep{evalplus}, and LiveCodeBench (v6 only, February 2025$\sim$May 2025)~\citep{lcb}. In all evaluations, we set the temperature to 1.0, top-p to 1.0, and the maximum generation length to 16,384. 
On each math reasoning benchmark, we sample 32 solutions for each problem; whereas each code generation benchmark, we sample 4 solutions per problem. We then report the average accuracy of each model on each benchmark. We adopt \texttt{Math-Verify}\footnote{\url{https://github.com/huggingface/Math-Verify}} as a rule-based verifier to validate answer correctness for math reasoning benchmarks.

\begin{figure}[t]  
\begin{center}
\begin{minipage}[t]{0.49\linewidth}
\centerline{\includegraphics[width=1\linewidth]{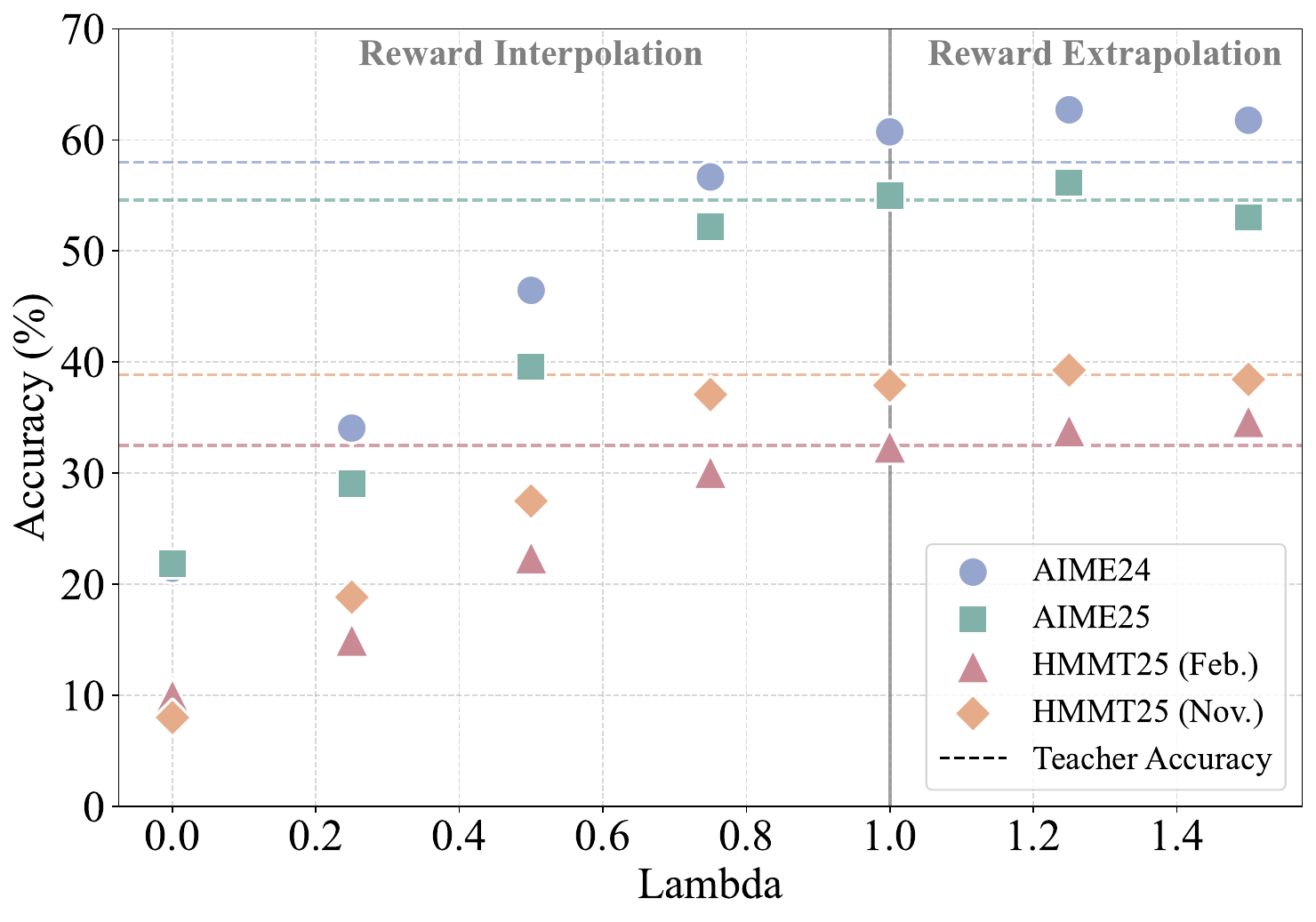}}
\caption{On-policy distillation results on four \textbf{math reasoning} benchmarks under different choices of reward scaling factor $\lambda$.}
\label{fig: qwen3-4b acc v.s. lambda math}
\end{minipage}  
\hfill
\begin{minipage}[t]{0.49\linewidth}
\centerline{\includegraphics[width=1\linewidth]{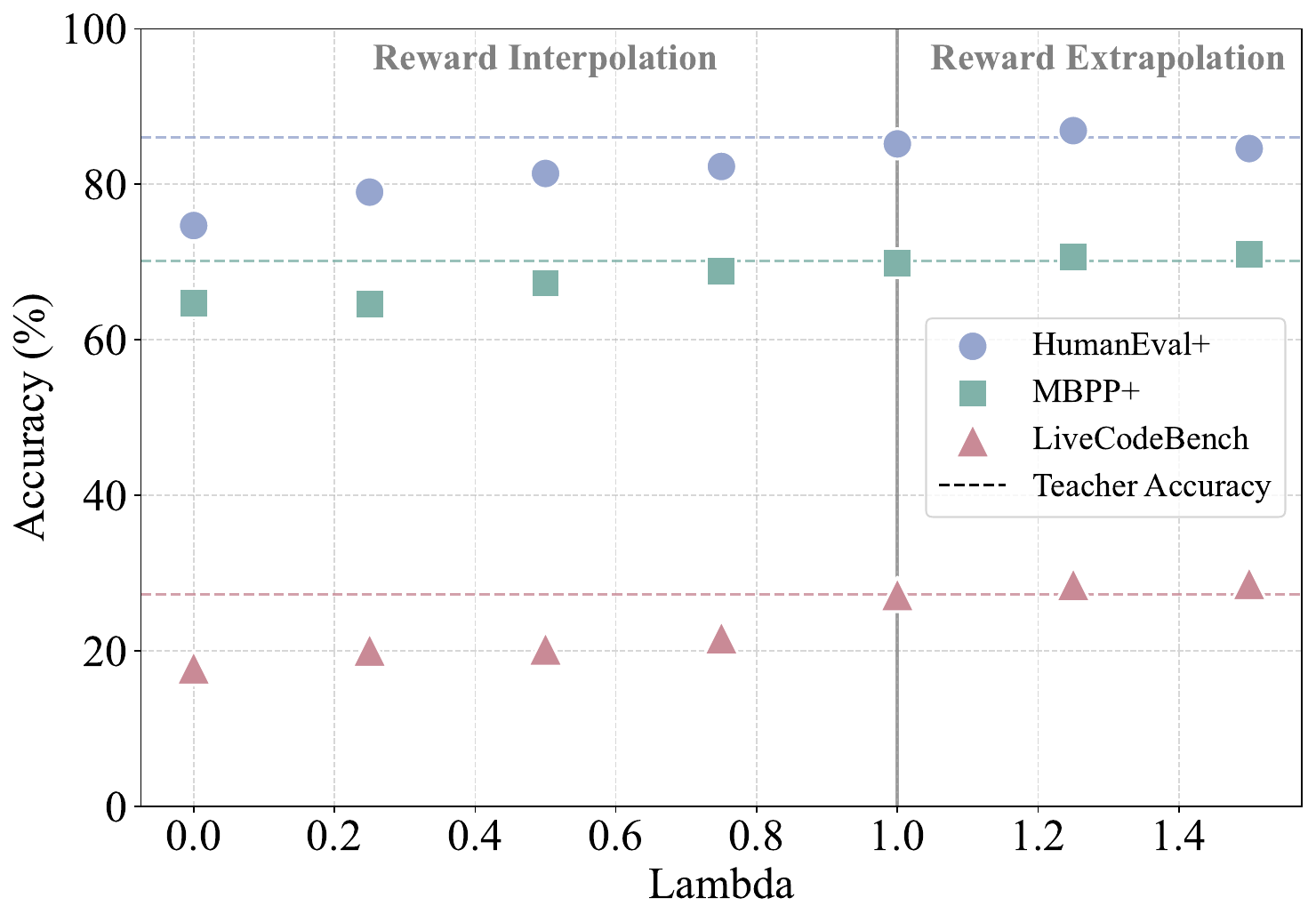}}
\caption{On-policy distillation results on three \textbf{code generation} benchmarks under different choices of reward scaling factor $\lambda$.}
\label{fig: qwen3-4b acc v.s. lambda code}
\end{minipage}  
\end{center}
\end{figure}

\begin{figure*}[t]
  \centering
  \subfigure[Results on AIME24 \label{fig: aime24 length vs acc}]{\includegraphics[width=0.32\textwidth]{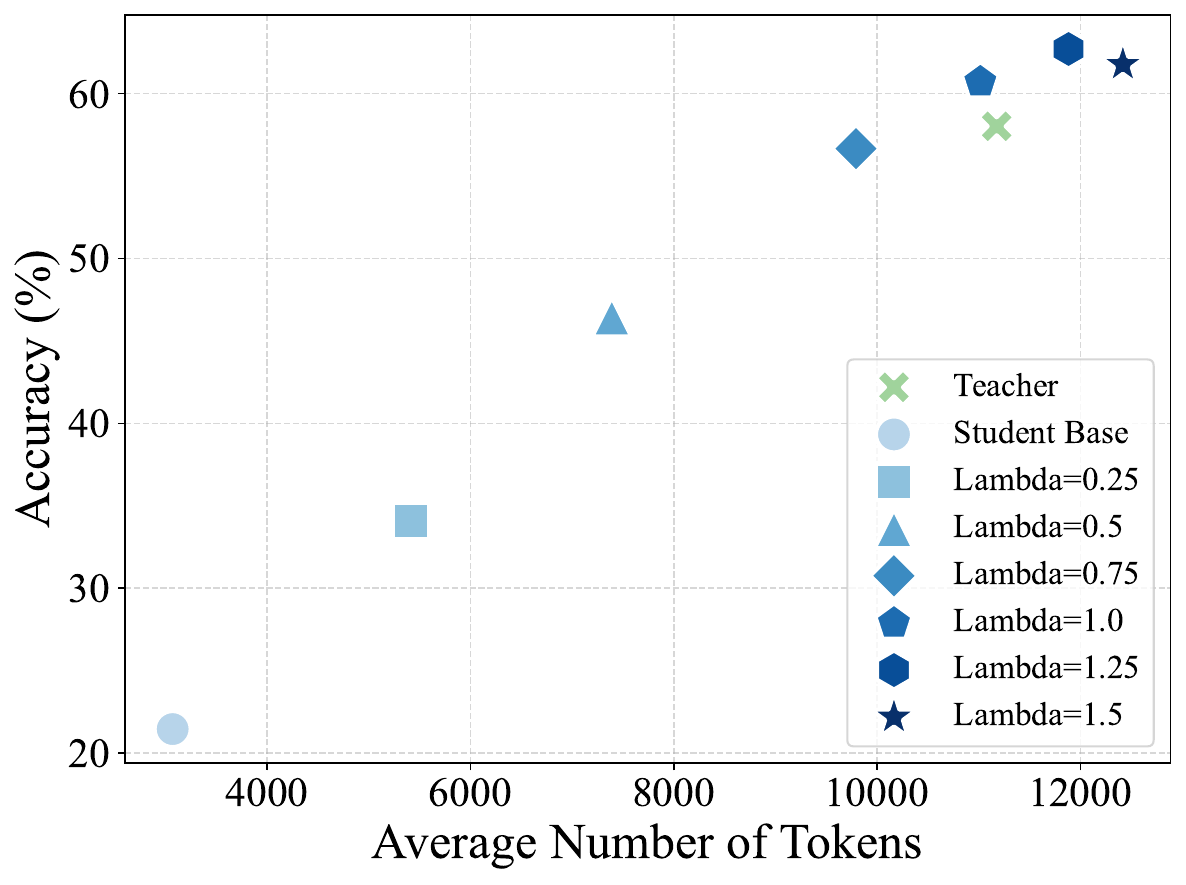}
  }
  \hfill
  \subfigure[Results on AIME25 \label{fig: aime25 length vs acc}]{\includegraphics[width=0.32\textwidth]{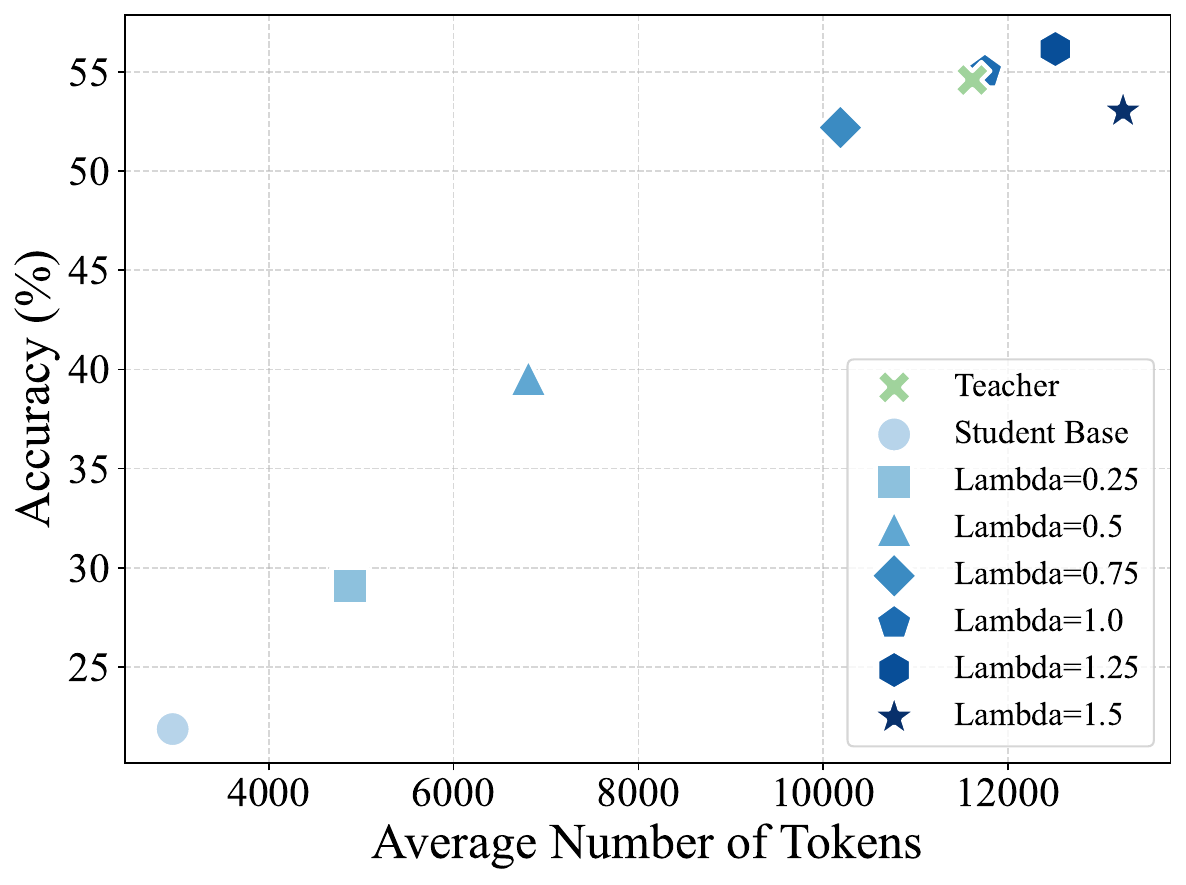}
  }
  \hfill
  \subfigure[Results on HMMT25 (Feb.) \label{fig: hmmt25 feb length vs acc}]{\includegraphics[width=0.32\textwidth]{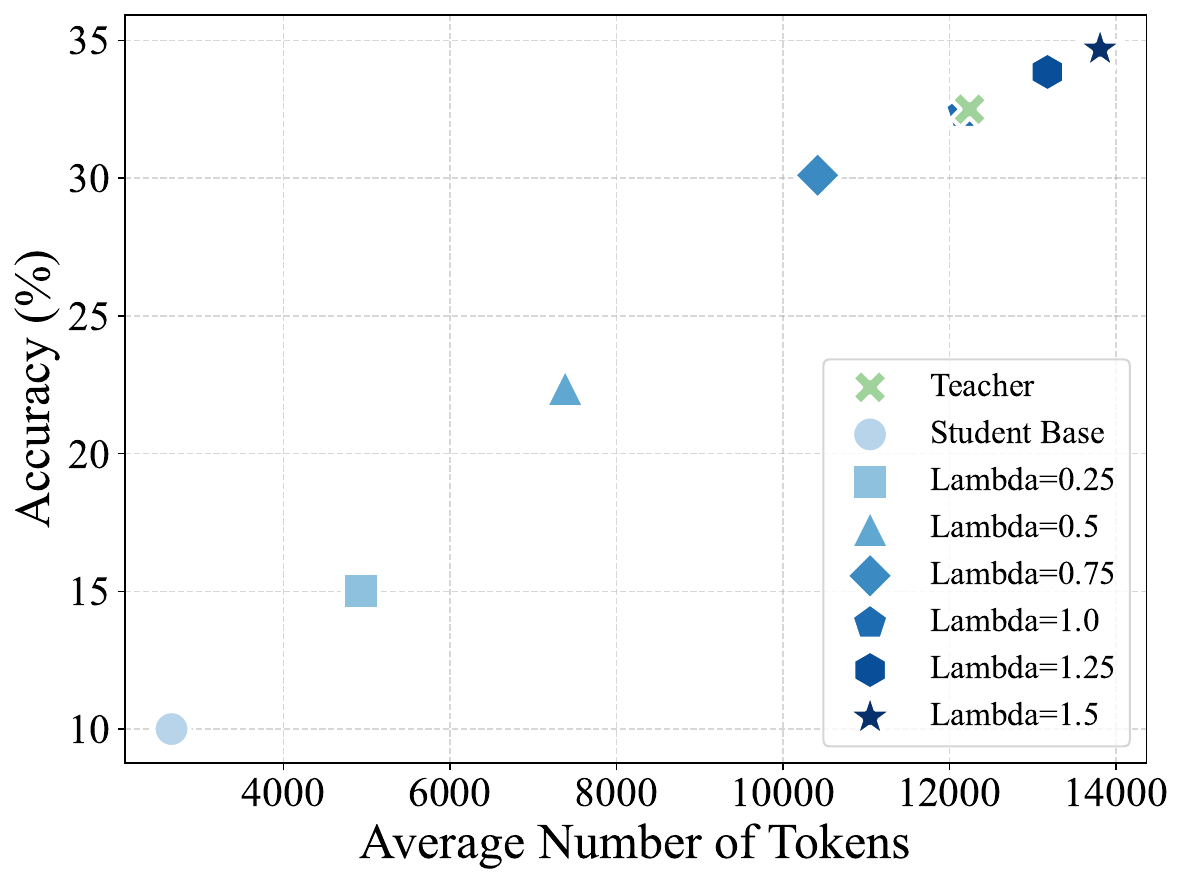}
  }
  \hfill
   \subfigure[Results on HumanEval+ \label{fig: humaneval+ length vs acc}]{\includegraphics[width=0.31\textwidth]{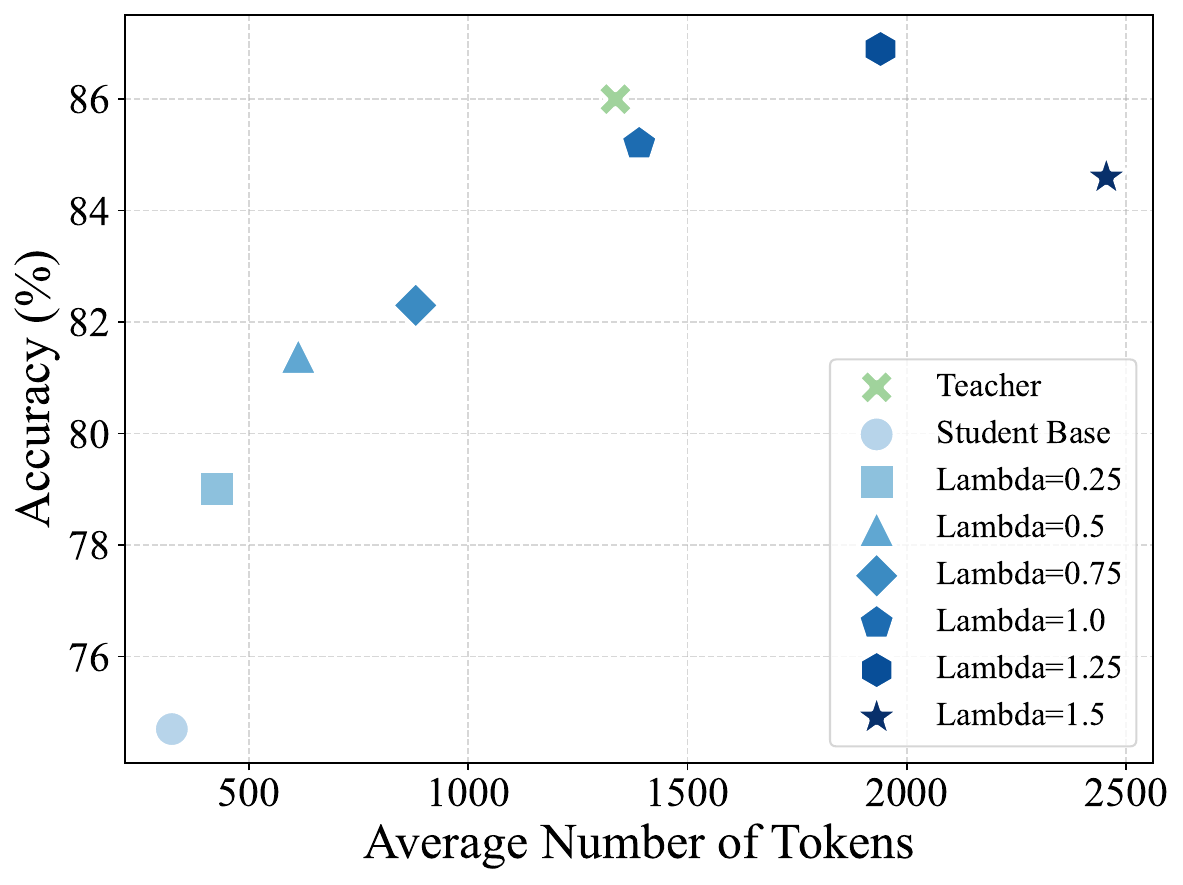}
  }
  \hfill
  \subfigure[Results on MBPP+ \label{fig: mbpp+ length vs acc}]{\includegraphics[width=0.31\textwidth]{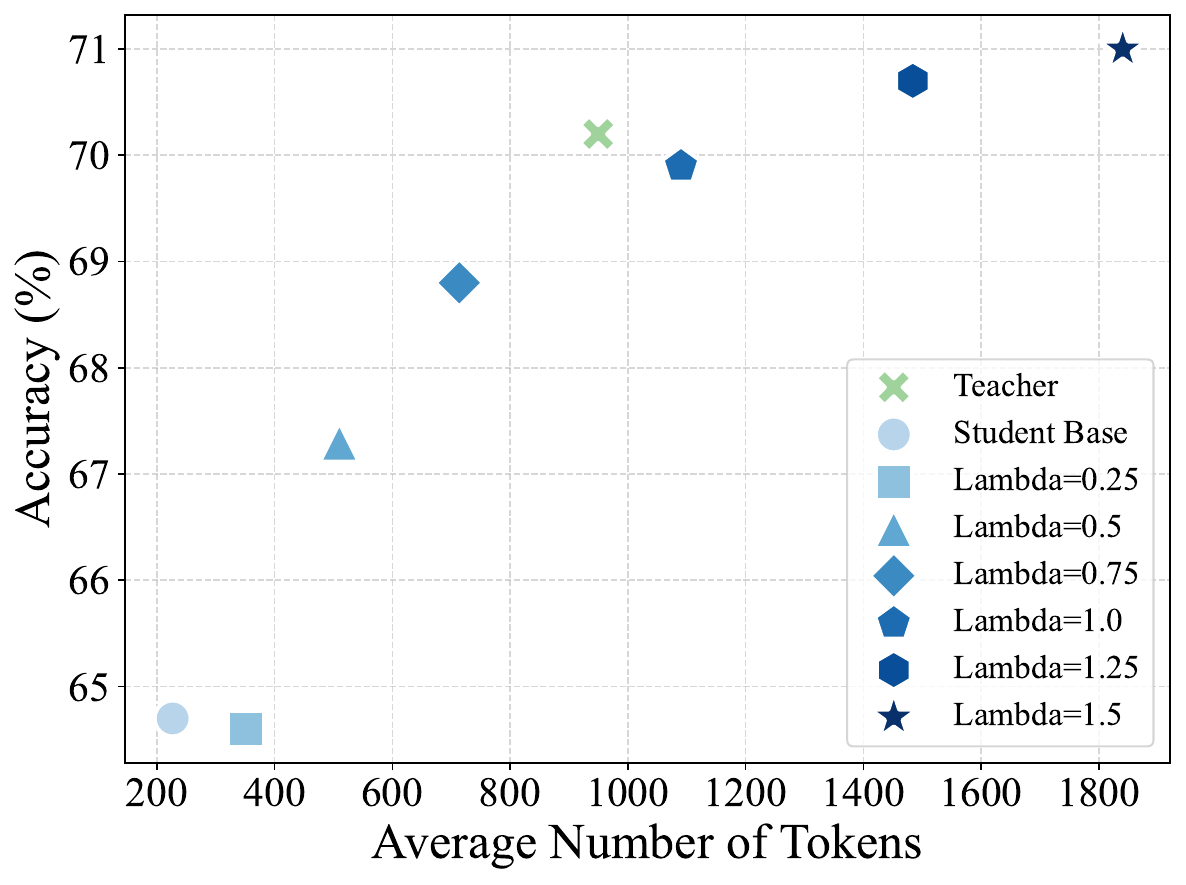}
  }
  \hfill
  \subfigure[Results on LiveCodeBench \label{fig: lcb length vs acc}]{\includegraphics[width=0.31\textwidth]{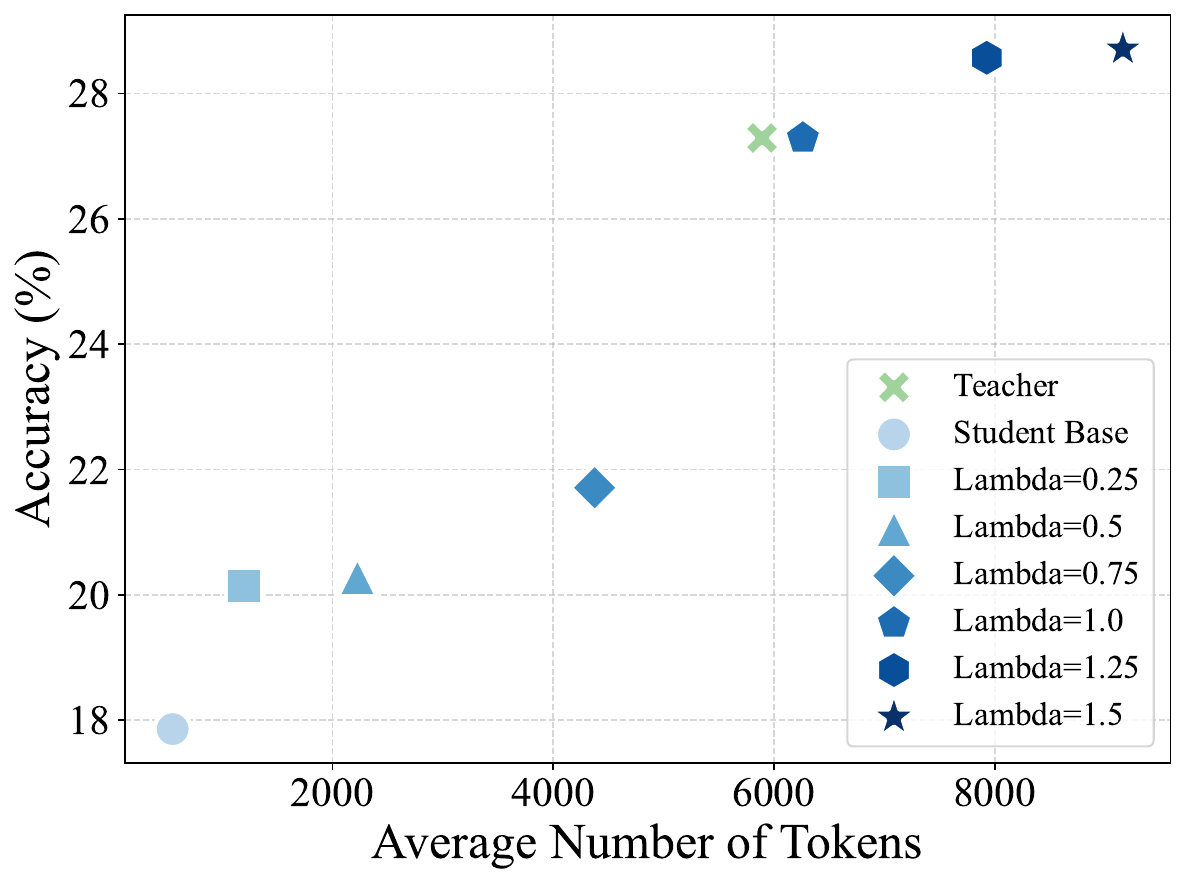}
  }
  \hfill
  \caption{Trends in the average number of tokens and the average accuracy of the on-policy distilled models across different benchmarks under varying reward scaling factors. The teacher for math reasoning tasks is Qwen3-4B-Non-Thinking-RL-Math, while the teacher for code generation tasks is Qwen3-4B-Non-Thinking-RL-Code.  
  }
  \label{fig: qwen3-4b length vs acc}
\end{figure*}

\subsubsection{Results of Single-Teacher Distillation}
\label{subsubsec: single-teacher distillation}
We first explore the impact of reward scaling factor $\lambda$ in G-OPD in the same-sized single-teacher distillation setting as the preliminary experiments (i.e., distilling Qwen3-4B-Non-Thinking-RL-Math or Qwen3-4B-Non-Thinking-RL-Code back into Qwen3-4B-Non-Thinking). The evaluation results in math reasoning and code generation domains are in Figure~\ref{fig: qwen3-4b acc v.s. lambda math} and Figure~\ref{fig: qwen3-4b acc v.s. lambda code} respectively.  We also visualize the relationship between accuracy and response length of each model in Figure~\ref{fig: qwen3-4b length vs acc} for deep analysis.

We can draw the following conclusions: (1) Standard OPD can fully recover the post-training behavior. As we can see, the student produced by OPD closely matches the evaluation accuracy and response length of the domain teacher. (2) Reward interpolation ($0 < \lambda < 1$) produces a student whose behavior (performance and response length) lies between the base model and the teacher model. Also, both the performance and response length increase monotonically as $\lambda$ grows, approaching the behavior of the teacher. This property can be leveraged to achieve budget-controlled reasoning~\citep{tops,orbit}. (3) \textbf{Reward extrapolation ($\lambda > 1$) outperforms standard OPD and has the potential to produce a student that surpasses the domain teacher.} As observed, ExOPD with appropriate reward extrapolation (i.e., $\lambda = 1.25$) consistently outperforms OPD and the domain teacher in all settings (also see Table~\ref{tab: results on same-sized models}), while excessive reward extrapolation (i.e., $\lambda = 1.5$) may lead to instability and degrade performance. This can be explained by the fact that continuously increasing $\lambda$ introduces the risk of the student hacking the implicit reward in Eq.~(\ref{eq:rewards_in_opd}), by aggressively fitting the peak of the log ratio, even if some tokens have excessively large log ratios due to bias. Furthermore, we can see that the response lengths of the students produced by ExOPD continue to increase, which may be due to the length bias issue of the implicit reward~\citep{laser}.

To demonstrate that the improvement of ExOPD over the teacher is not due to less training of the teacher, we compare the evaluation performance of ExOPD and the teacher after an additional 100 steps of RL training. The results in Table~\ref{tab: comparison with teacher with additional training} show that the teacher with more continued RL training show smaller improvement compared to ExOPD with fewer steps. We also demonstrate the generalizability and effectiveness of ExOPD when the teacher models are trained with sufficient RL steps (i.e., 1200 RL steps), with the corresponding results provided in Appendix~\ref{appendix: results under stronger teachers}.~\looseness=-1

\input{tables/comparison_against_teacher_with_additional_rl}

\subsubsection{Results of Multi-Teacher Distillation}
\label{subsubsec: multi-teacher distillation}
Based on above analysis, we conduct experiments in the multi-teacher distillation setting, where we aim to merge the capabilities from different domain teachers, obtained by applying domain-specific RL to the same base model, into the original base model through OPD~\citep{mimo}. This has been demonstrated to be an effective new multi-task post-training paradigm. Specifically, the domain teachers are the above RL variants Qwen3-4B-Non-Thinking-RL-Math/Code, and the student model is Qwen3-4B-Non-Thinking. From the preliminary results in Section~\ref{subsubsec: single-teacher distillation}, we can see that $\lambda=1.25$ in ExOPD consistently leads to better performance than OPD. Thus, \textbf{in all subsequent experiments, we fix $\lambda = 1.25$ for ExOPD without any further specific tuning}. 
Besides OPD, we also compare against two baselines: (1) \textbf{Supervised fine-tuning} (\textbf{SFT}), which trains the student on the teachers' generated trajectories via Cross-Entropy Loss. We ensure that the number of trajectories used for SFT is consistent with those in OPD and ExOPD. More details can be found in Appendix~\ref{appendix: training settings}. (2) \textbf{ExPO}~\citep{expo}, a weight extrapolation method. We implement ExPO by first averaging the weights of all domain teachers, then extrapolating the weights against the student model using an extrapolation factor $\alpha$, which is tuned from $\{0.25, 0.5\}$ following the recommendations. For a fair comparison, we downweight the sample size of the math RL data to match that of the code RL data in both OPD and ExOPD here, ensuring that each domain has the same sample size.

\input{tables/results_on_same_size_models}

The results of multi-teacher distillation are shown in Table~\ref{tab: results on same-sized models}. As we can see, SFT produces a sub-optimal student, while the performance ceiling of OPD is typically bounded by the teachers. ExPO, though training-free, cannot ensure that the weight-extrapolated student consistently surpasses all domain teachers, lacking good controllability. 
However, \textbf{our method ExOPD consistently outperforms OPD and is the only method that produces a unified student capable of surpassing both domain teachers on all benchmarks}. 

\begin{figure*}[t]
  \centering
  \begin{tabular}{c}
\includegraphics[width=0.95\linewidth]{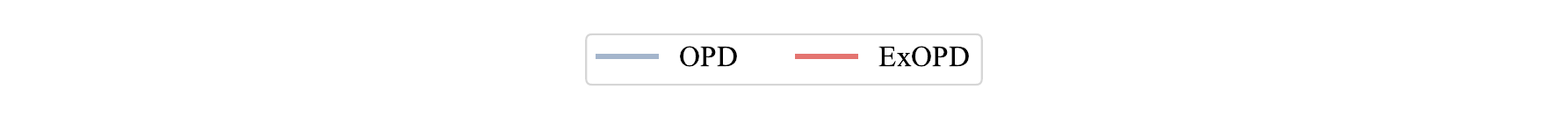}
\end{tabular}
\vskip -0.2in
  \subfigure[Training rewards \label{fig: multi-teacher distillation rewards}]{\includegraphics[width=0.32\textwidth]{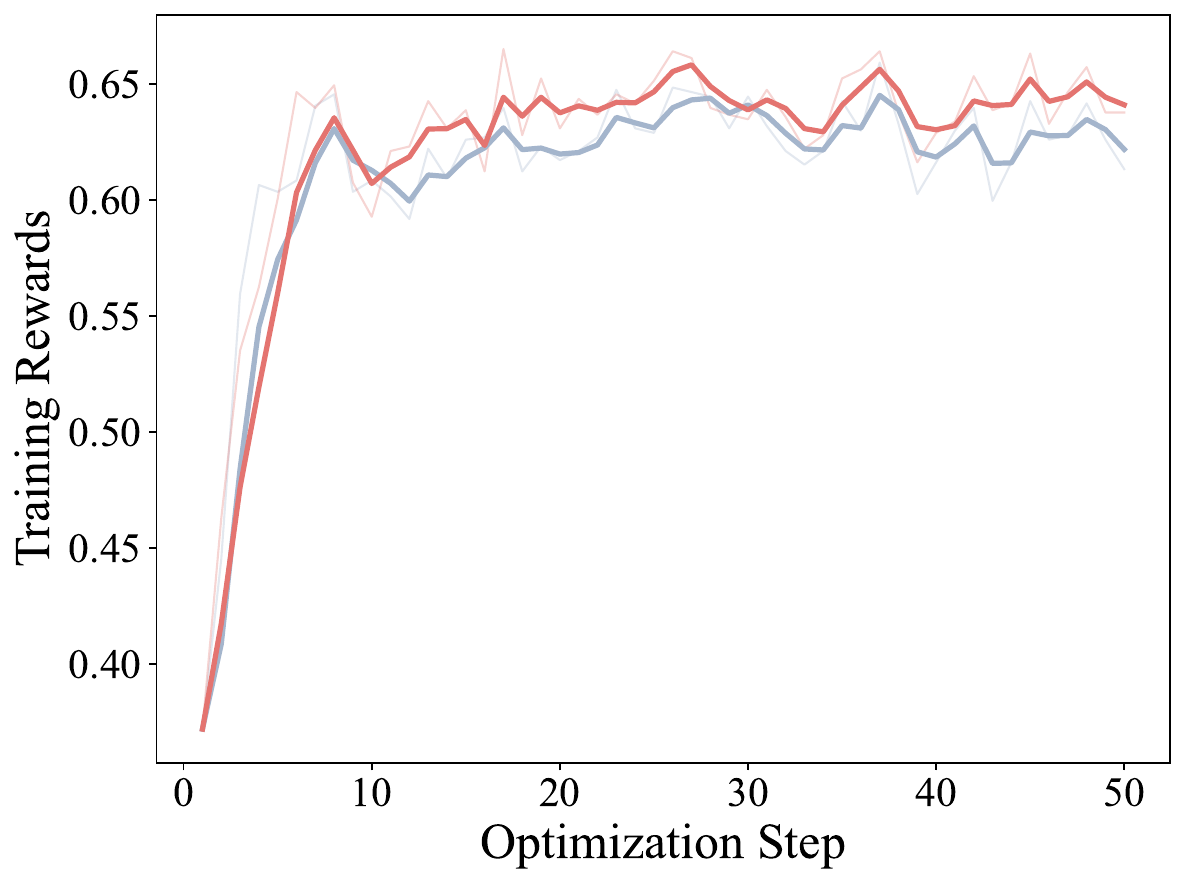}
  }
  \hfill
  \subfigure[Response length \label{fig: multi-teacher distillation length}]{\includegraphics[width=0.32\textwidth]{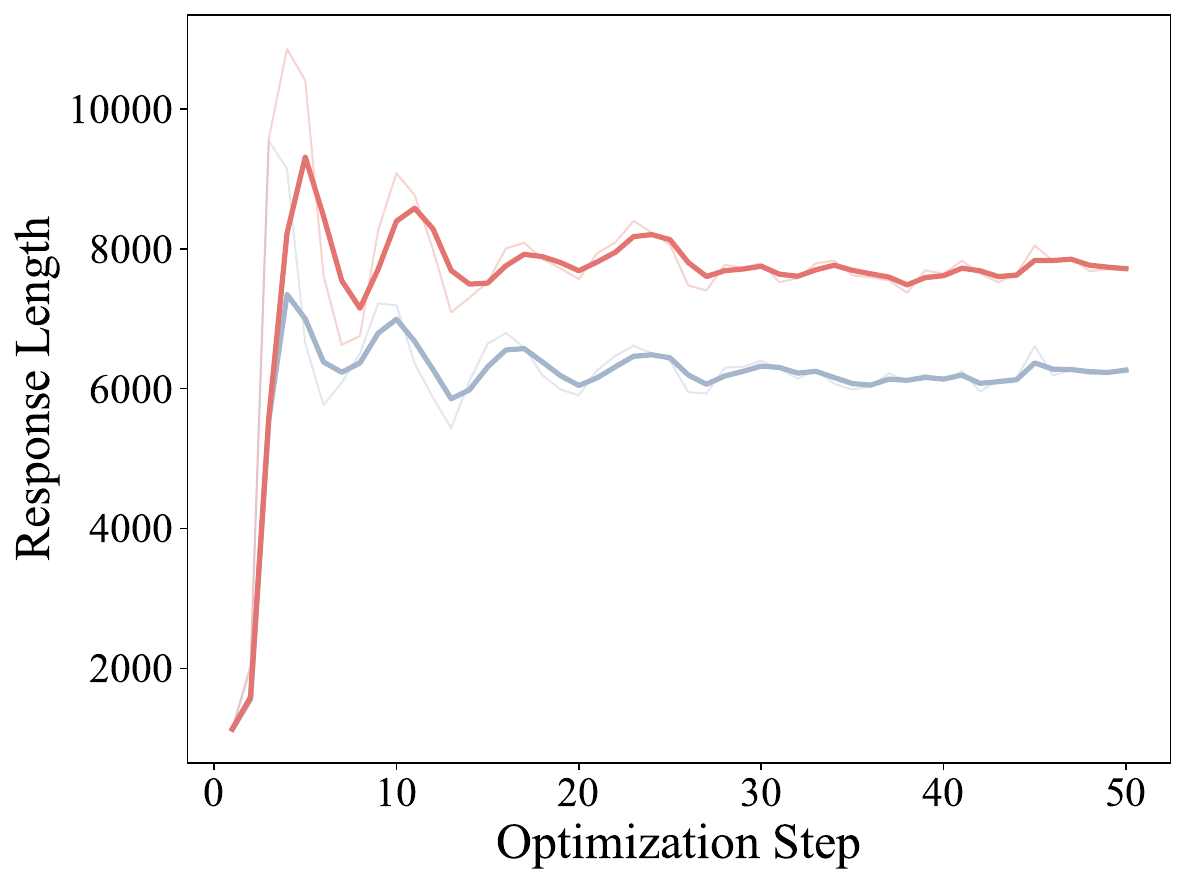}
  }
  \hfill
  \subfigure[Entropy \label{fig: multi-teacher distillation entropy}]{\includegraphics[width=0.32\textwidth]{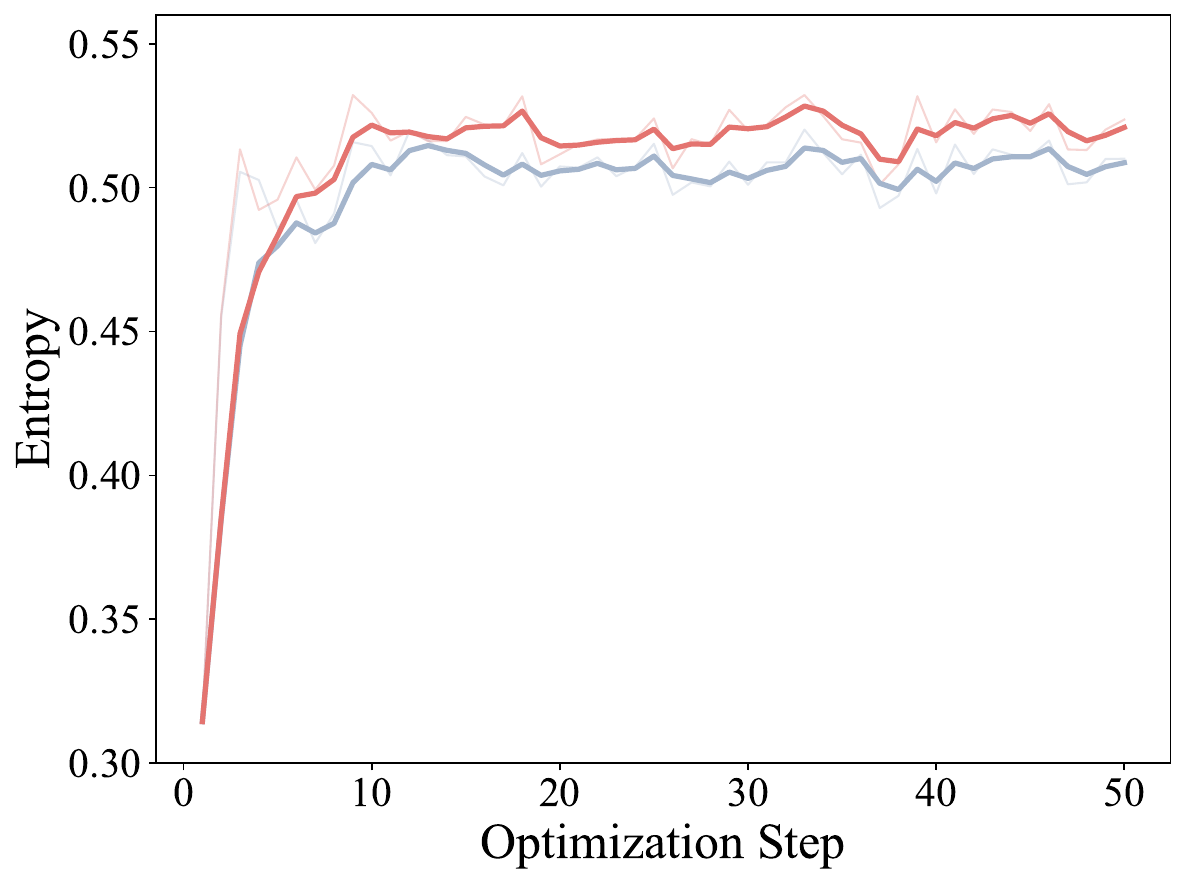}
  }
  \caption{Training dynamics of OPD and ExOPD in multi-teacher distillation experiments. We visualize using Exponential Moving Average (EMA) smoothing with a coefficient of 0.5.
  }
  \label{fig: training dynamics}
  \vskip -0.05in
\end{figure*}

Furthermore, we analyze the training dynamics of ExOPD compared to OPD to gain a deeper understanding of ExOPD. We put the comparison in Figure~\ref{fig: training dynamics}. ExOPD achieves higher training rewards but makes the student generate longer response lengths, which is consistent with the evaluation results shown in Figure~\ref{fig: qwen3-4b length vs acc}. We also observe that the response entropy of the student trained by ExOPD is higher than that trained by OPD. We attribute this to the fact that the former tends to generate longer responses, increasing the response diversity.

\subsection{Experiments in the Strong-to-Weak Distillation Setting}
\label{subsec: strong-to-weak distillation}
Another practical usage of OPD is for strong-to-weak distillation~\citep{qwen3}, i.e., distilling capabilities from a larger teacher into a smaller student. Thus, in this section, we explore the effectiveness of ExOPD and the additional reward correction practice in the strong-to-weak distillation setting. 

\subsubsection{Experimental Settings}
We select Qwen3-30B-A3B-Instruct-2507 as the teacher model and perform distillation on Qwen3-1.7B-Non-Thinking and Qwen3-4B-Non-Thinking, respectively. We primarily conduct experiments in the math reasoning domain, where the training and evaluation datasets are the same as those used in Section~\ref{subsec: same-sized distillation}. The training details are in Appendix~\ref{appendix: training settings}. In ExOPD, we first conduct experiments in the default setting (Section~\ref{subsubsec: strong-to-weak results}), where we assume the availability of only two models: the student base model and the stronger teacher model. Thus, in this default setting, we set the reference model in ExOPD to the student base model. We also explore the effectiveness of the reward correction technique in Section~\ref{subsubsec: reward correction}, where we assume extra access to the teacher's pre-RL variant, which serves as the reference model in ExOPD. We compare ExOPD against standard OPD and off-policy distillation (SFT). 

\input{tables/results_on_different_sized_models}

\subsubsection{Results of Strong-to-Weak Distillation}
\label{subsubsec: strong-to-weak results}
The results in default strong-to-weak distillation setting are put in Table~\ref{tab: math results on different-sized models}. The main conclusion is that \textbf{ExOPD can bring significant improvements in strong-to-weak distillation, outperforming off-policy distillation and standard OPD by a large margin}. The results reveal that, although the implicit reward $\log \frac{\pi^*}{\pi_{\text{base}}^{\text{student}}}$ may contain noise due to the intrinsic knowledge gap and distribution bias between the small and large models, extrapolating the rewards can still push the limits of OPD in strong-to-weak distillation.

\subsubsection{Reward Correction in Strong-to-Weak Distillation}
\label{subsubsec: reward correction}
As shown above, the default ExOPD with the reference model fixed as the student base model can already bring significant improvement over OPD. However, as discussed in Remark~\ref{remark: G-OPD}, setting the reference model to the teacher's pre-RL variant—if available—may further enhance the distillation performance. Here, we conduct experiments to validate this analysis. Specifically, since we cannot get the pre-RL variant of Qwen3-30B-A3B-Instruct-2507, we choose our trained Qwen3-4B-Non-Thinking-RL-Math/Code as the teachers and take Qwen3-4B-Non-Thinking as the pre-RL variant. The student model is Qwen3-1.7B-Non-Thinking.

\begin{figure*}[t]
  \centering
  \subfigure[Results averaged on four \textbf{math reasoning} benchmarks, student is Qwen3-1.7B-Non-Thinking, teacher is Qwen3-4B-Non-Thinking-RL-Math \label{fig: qwen3-1.7b reward correction math}]{\includegraphics[width=0.48\textwidth]{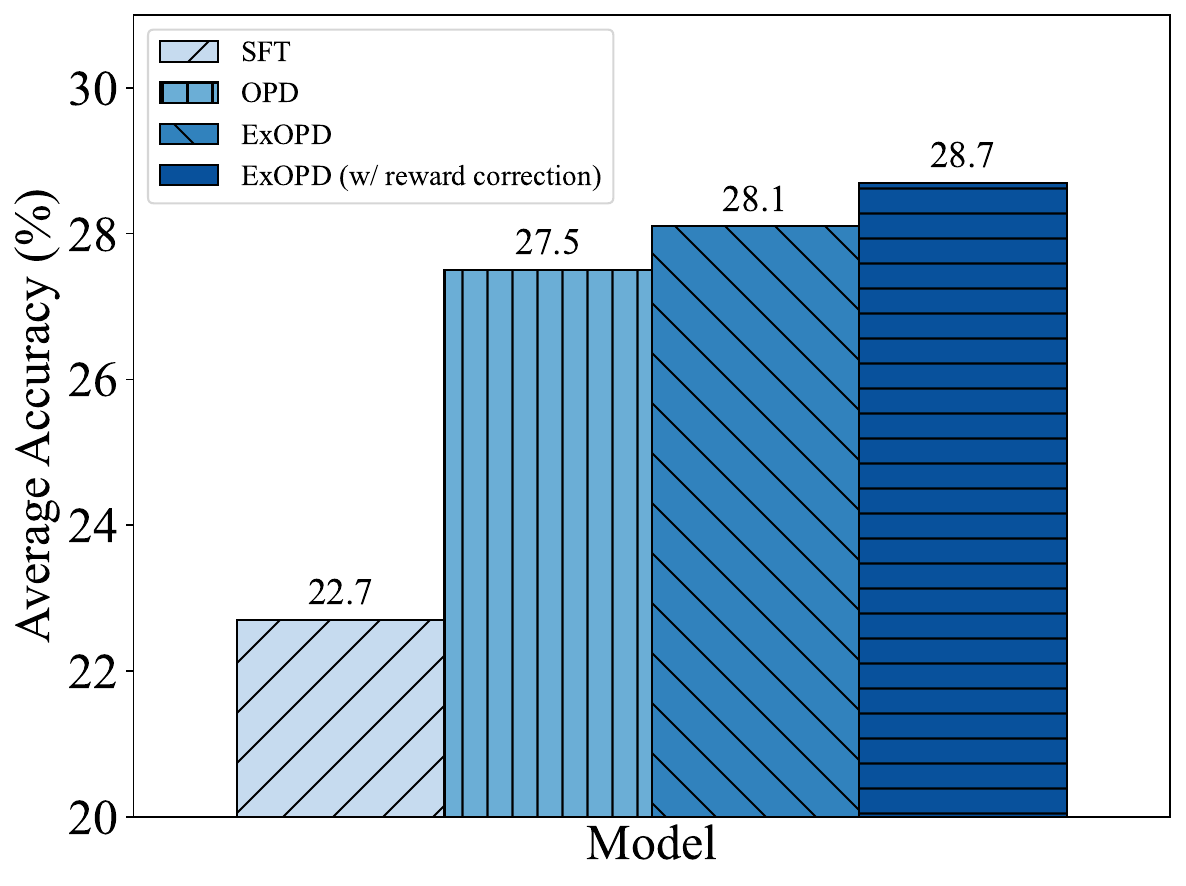}
  }
  \hfill
  \subfigure[Results averaged on three \textbf{code generation} benchmarks, student is Qwen3-1.7B-Non-Thinking, teacher is Qwen3-4B-Non-Thinking-RL-Code \label{fig: qwen3-1.7b reward correction code}]{\includegraphics[width=0.48\textwidth]{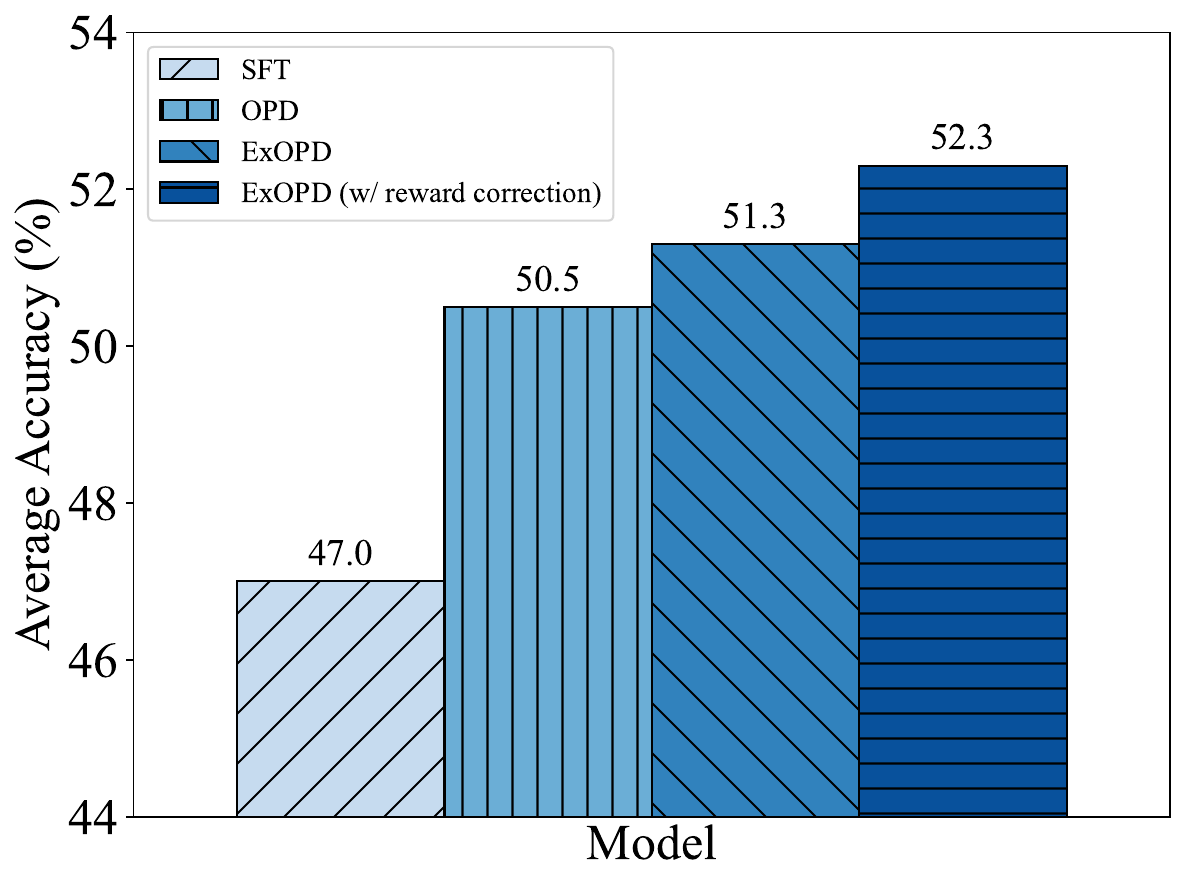}
  }
  \caption{Effect of \textit{reward correction} in the strong-to-weak distillation setting.}
  \label{fig: effect of reward correction}
  \vskip -0.1in
\end{figure*}

The comparison results are displayed in Figure~\ref{fig: effect of reward correction}. The results validate the effectivenss of the reward correction practice, which \textbf{consistently boosts the performance of ExOPD}. However, we reiterate that reward correction requires access to $\pi_{\text{base}}^{\text{teacher}}$ and incurs higher computational cost, since it requires computing log-probabilities under a larger reference model than in the default ExOPD.

\section{Conclusion and Discussion}
In this work, we conduct an in-depth analysis of the on-policy distillation paradigm. We first establish an interesting connection between OPD and dense KL-constrained RL. Building on this insight, we propose a generalized OPD framework (G-OPD) by introducing (i) a flexible reference model for the implicit reward function and (ii) a reward scaling factor that controls the relative weight of the reward term versus KL regularization. Through comprehensive experiments on math reasoning and code generation tasks, we provide several novel insights: (1) Appropriate reward extrapolation (i.e., setting the reward scaling factor to be larger than 1) can improve OPD performance, and in same-sized multi-teacher distillation it enables learning a unified student that surpasses all domain-specific teachers. We refer to this variant as ExOPD. (2) Moreover, in strong-to-weak distillation, replacing the student's initial policy with the teacher's pre-RL policy as the reference model can further boost the performance of ExOPD. 

Regarding future work, we believe it is practical to explore: (1) validating the generalizability of ExOPD on larger-scale models; (2) assessing the robustness of ExOPD in multi-teacher distillation with a broader and more diverse set of domain teachers; and (3) evaluating the effectiveness of ExOPD for on-policy distillation across different model families.

\newpage 
\bibliography{iclr2025_conference}
\bibliographystyle{iclr2025_conference}

\clearpage
\appendix

\section{Detailed Math Derivations}
\label{appendix: math derivations}
Here, we make mathematical derivations to calculate the expected gradients of OPD objective in Eq.~(\ref{eq:opd}).

Since 
\begin{equation}
\begin{aligned}
\mathcal{J}_{\text{OPD}}(\boldsymbol{\theta}) &=  \min_{\boldsymbol{\theta}}\;
\mathbb{E}_{\boldsymbol{x} \sim D, \boldsymbol{y} \sim \pi_{\boldsymbol{\theta}} (\cdot | \boldsymbol{x}) }\Big[
\mathcal{D}_{\mathrm{KL}} \!\big(
\pi_{\boldsymbol{\theta}}(\boldsymbol{y} | \boldsymbol{x})
\,\big\|\,
\pi^{*}( \boldsymbol{y} | \boldsymbol{x})
\big)
\Big] \\ & = \min_{\boldsymbol{\theta}}\;
\mathbb{E}_{\boldsymbol{x} \sim D, \boldsymbol{y} \sim \pi_{\boldsymbol{\theta}} (\cdot | \boldsymbol{x}) }\Big[
\log \pi_{\boldsymbol{\theta}}(\boldsymbol{y}| \boldsymbol{x} ) - \log \pi^{*}(\boldsymbol{y}| \boldsymbol{x} )
\Big].
\end{aligned}
\end{equation}
We can get
\begin{equation}
\begin{aligned}
\label{eq: derive opd grad}
\nabla_{\boldsymbol{\theta}} \mathcal{J}_{\text{OPD}}(\boldsymbol{\theta}) &=  
\nabla_{\boldsymbol{\theta}}  \mathbb{E}_{\boldsymbol{x} \sim D, \boldsymbol{y} \sim \pi_{\boldsymbol{\theta}} (\cdot | \boldsymbol{x}) }\Big[
\log \pi_{\boldsymbol{\theta}}(\boldsymbol{y}| \boldsymbol{x} ) - \log \pi^{*}(\boldsymbol{y}| \boldsymbol{x} )
\Big] \\ &=
\nabla_{\boldsymbol{\theta}}  \mathbb{E}_{\boldsymbol{x}} \Big[ \sum_{\boldsymbol{y}} \pi_{\boldsymbol{\theta}} (\boldsymbol{y}|\boldsymbol{x})\big(  
\log \pi_{\boldsymbol{\theta}}(\boldsymbol{y}| \boldsymbol{x} ) - \log \pi^{*}(\boldsymbol{y}| \boldsymbol{x} )
\big)
\Big] \\ &=
\mathbb{E}_{\boldsymbol{x}}  \Big[ \nabla_{\boldsymbol{\theta}}  \sum_{\boldsymbol{y}} \pi_{\boldsymbol{\theta}} (\boldsymbol{y}|\boldsymbol{x})\big(  
\log \pi_{\boldsymbol{\theta}}(\boldsymbol{y}| \boldsymbol{x} ) - \log \pi^{*}(\boldsymbol{y}| \boldsymbol{x} )
\big)
\Big] \\ &=
\mathbb{E}_{\boldsymbol{x}}  \Big[ \sum_{\boldsymbol{y}} \big(\nabla_{\boldsymbol{\theta}}  \pi_{\boldsymbol{\theta}} (\boldsymbol{y}|\boldsymbol{x}) \big) \big(  
\log \pi_{\boldsymbol{\theta}}(\boldsymbol{y}| \boldsymbol{x} ) - \log \pi^{*}(\boldsymbol{y}| \boldsymbol{x} )
\big) + \sum_{\boldsymbol{y}} \pi_{\boldsymbol{\theta}}  (\boldsymbol{y}| \boldsymbol{x} ) \nabla_{\boldsymbol{\theta}}   \log \pi_{\boldsymbol{\theta}}(\boldsymbol{y}| \boldsymbol{x} )
\Big]  .
\end{aligned}
\end{equation}
Notice that
\begin{equation}
\begin{aligned}
\mathbb{E}_{\boldsymbol{x}}  \Big[ 
\sum_{\boldsymbol{y}} \pi_{\boldsymbol{\theta}}  (\boldsymbol{y}| \boldsymbol{x} ) \nabla_{\boldsymbol{\theta}}   \log \pi_{\boldsymbol{\theta}}(\boldsymbol{y}| \boldsymbol{x} )
\Big] & = \mathbb{E}_{\boldsymbol{x}}  \Big[ 
\sum_{\boldsymbol{y}} \pi_{\boldsymbol{\theta}}  (\boldsymbol{y}| \boldsymbol{x} ) \frac{ \nabla_{\boldsymbol{\theta}} \pi_{\boldsymbol{\theta}}  (\boldsymbol{y}| \boldsymbol{x} )}{\pi_{\boldsymbol{\theta}}  (\boldsymbol{y}| \boldsymbol{x} )}
\Big] \\ &= 
\mathbb{E}_{\boldsymbol{x}} \Big[ 
\sum_{\boldsymbol{y}} \nabla_{\boldsymbol{\theta}} \pi_{\boldsymbol{\theta}}  (\boldsymbol{y}| \boldsymbol{x} )
\Big] \\ & =
\mathbb{E}_{\boldsymbol{x}}  \Big[
\nabla_{\boldsymbol{\theta}} \sum_{\boldsymbol{y}}  \pi_{\boldsymbol{\theta}}  (\boldsymbol{y}| \boldsymbol{x} )
\Big] \\ &=
\mathbb{E}_{\boldsymbol{x}}  \Big[
\nabla_{\boldsymbol{\theta}}  1
\Big] \\ &= 0.
\end{aligned}
\end{equation}
Therefore, Eq.~(\ref{eq: derive opd grad}) can be reduced to
\begin{equation}
\begin{aligned}
\label{eq:opd_grad_2}
\nabla_{\boldsymbol{\theta}} \mathcal{J}_{\text{OPD}}(\boldsymbol{\theta}) &= \mathbb{E}_{\boldsymbol{x}}  \Big[ \sum_{\boldsymbol{y}}  \nabla_{\boldsymbol{\theta}}  \pi_{\boldsymbol{\theta}} (\boldsymbol{y}|\boldsymbol{x}) \big(  
\log \pi_{\boldsymbol{\theta}}(\boldsymbol{y}| \boldsymbol{x} ) - \log \pi^{*}(\boldsymbol{y}| \boldsymbol{x} )
\big)
\Big]  \\ &=
\mathbb{E}_{\boldsymbol{x}}  \Big[ \sum_{\boldsymbol{y}} 
\pi_{\boldsymbol{\theta}} (\boldsymbol{y}|\boldsymbol{x}) \nabla_{\boldsymbol{\theta}}  \log \pi_{\boldsymbol{\theta}} (\boldsymbol{y}|\boldsymbol{x})\big(  
\log \pi_{\boldsymbol{\theta}}(\boldsymbol{y}| \boldsymbol{x} ) - \log \pi^{*}(\boldsymbol{y}| \boldsymbol{x} )
\big)
\Big]
\\ &= 
\mathbb{E}_{\boldsymbol{x} \sim D, \boldsymbol{y} \sim \pi_{\boldsymbol{\theta}} (\cdot | \boldsymbol{x})}  \Big[ \big(  
\log \pi_{\boldsymbol{\theta}}(\boldsymbol{y}| \boldsymbol{x} ) - \log \pi^{*}(\boldsymbol{y}| \boldsymbol{x} )
\big) \nabla_{\boldsymbol{\theta}}  \log \pi_{\boldsymbol{\theta}} (\boldsymbol{y}|\boldsymbol{x})
\Big] \\ & =
\mathbb{E}_{\boldsymbol{x} \sim D, \boldsymbol{y} \sim \pi_{\boldsymbol{\theta}} (\cdot | \boldsymbol{x}) }\Big[
\sum_{t=1}^{T}  \sum_{t^{'} =1}^{T} \big(\log \pi_{\boldsymbol{\theta}}(y_{t^{'}}| \boldsymbol{x}, \boldsymbol{y}_{<t^{'}} ) - \log \pi^{*}(y_{t^{'}}| \boldsymbol{x}, \boldsymbol{y}_{<t^{'}} ) \big) \, \nabla_{\boldsymbol{\theta}} \log \pi_{\boldsymbol{\theta}}(y_t| \boldsymbol{x}, \boldsymbol{y}_{<t} ) 
\Big]. 
\end{aligned}
\end{equation}
Now let's denote 
$$\Delta_{t^{'}}=\big(\log \pi_{\boldsymbol{\theta}}(y_{t^{'}}| \boldsymbol{x}, \boldsymbol{y}_{<t^{'}} ) - \log \pi^{*}(y_{t^{'}}| \boldsymbol{x}, \boldsymbol{y}_{<t^{'}} ) \big),
$$ and consider each term $\mathbb{E}_{ \boldsymbol{x} \sim D,\boldsymbol{y} \sim \pi_{\boldsymbol{\theta}} (\cdot | \boldsymbol{x}) } \Big[ \Delta_{t^{'}} \, \nabla_{\boldsymbol{\theta}} \log \pi_{\boldsymbol{\theta}}(y_t| \boldsymbol{x}, \boldsymbol{y}_{<t} ) \Big] $ where $t^{'} < t$:
\begin{equation}
\begin{aligned}
\mathbb{E}_{\boldsymbol{x},\boldsymbol{y}} \Big[ \Delta_{t^{'}} \, \nabla_{\boldsymbol{\theta}} \log \pi_{\boldsymbol{\theta}}(y_t| \boldsymbol{x}, \boldsymbol{y}_{<t} ) \Big] &= \mathbb{E}_{\boldsymbol{x},\boldsymbol{y}}  \Big[ \mathbb{E}_{y_{t}} \Big[   \Delta_{t^{'}} \, \nabla_{\boldsymbol{\theta}} \log \pi_{\boldsymbol{\theta}}(y_t| \boldsymbol{x}, \boldsymbol{y}_{<t} ) \big| \boldsymbol{x}, \boldsymbol{y}_{<t} \Big]\Big] \\ &=
\mathbb{E}_{\boldsymbol{x},\boldsymbol{y}}  \Big[ \Delta_{t^{'}} \mathbb{E}_{y_t} \Big[   \nabla_{\boldsymbol{\theta}} \log \pi_{\boldsymbol{\theta}}(y_t| \boldsymbol{x}, \boldsymbol{y}_{<t} ) \big| \boldsymbol{x}, \boldsymbol{y}_{<t} \Big]\Big] \\ &=
\mathbb{E}_{\boldsymbol{x},\boldsymbol{y}} \Big[ \Delta_{t^{'}} \mathbb{E}_{y_{t} \sim \pi_{\boldsymbol{\theta}}(\cdot|\boldsymbol{x},\boldsymbol{y}_{<t})} \big[   \nabla_{\boldsymbol{\theta}} \log \pi_{\boldsymbol{\theta}}(y_t| \boldsymbol{x}, \boldsymbol{y}_{<t} )  \big]\Big] \\ &=
\mathbb{E}_{\boldsymbol{x},\boldsymbol{y}} \Big[ \Delta_{t^{'}} \sum_{y_{t}}  \nabla_{\boldsymbol{\theta}}  \pi_{\boldsymbol{\theta}}(y_t| \boldsymbol{x}, \boldsymbol{y}_{<t} )  \Big] \\ &=
\mathbb{E}_{\boldsymbol{x},\boldsymbol{y}} \Big[ \Delta_{t^{'}}  \nabla_{\boldsymbol{\theta}}  \sum_{y_{t}}   \pi_{\boldsymbol{\theta}}(y_t| \boldsymbol{x}, \boldsymbol{y}_{<t} )  \Big] 
\\ &= 
\mathbb{E}_{\boldsymbol{x},\boldsymbol{y}} \Big[ \Delta_{t^{'}}  \nabla_{\boldsymbol{\theta}}  1 \Big]  
\\ &= 0.
\end{aligned} 
\end{equation}
Therefore, Eq.~(\ref{eq:opd_grad_2}) can be reduced to
\begin{equation}
\begin{aligned}
\label{eq:opd_grad_23}
\nabla_{\boldsymbol{\theta}} \mathcal{J}_{\text{OPD}}(\boldsymbol{\theta}) 
 &=
\mathbb{E}_{\boldsymbol{x} \sim D, \boldsymbol{y} \sim \pi_{\boldsymbol{\theta}} (\cdot | \boldsymbol{x}) }\Big[
\sum_{t=1}^{T} \Big( \sum_{t^{'} =t}^{T} \big(\log \pi_{\boldsymbol{\theta}}(y_{t^{'}}| \boldsymbol{x}, \boldsymbol{y}_{<t^{'}} ) - \log \pi^{*}(y_{t^{'}}| \boldsymbol{x}, \boldsymbol{y}_{<t^{'}} ) \big) \Big) \, \nabla_{\boldsymbol{\theta}} \log \pi_{\boldsymbol{\theta}}(y_t| \boldsymbol{x}, \boldsymbol{y}_{<t} ) 
\Big].
\end{aligned}
\end{equation}
In practice, recent studies~\citep{opd-tml,mimo} use a discount factor of 0 and approximate the gradient as
\begin{equation}
\begin{aligned}
\nabla_{\boldsymbol{\theta}} \mathcal{J}_{\text{OPD}}(\boldsymbol{\theta}) = 
\mathbb{E}_{\boldsymbol{x} \sim D, \boldsymbol{y} \sim \pi_{\boldsymbol{\theta}} (\cdot | \boldsymbol{x}) }\Big[
\sum_{t=1}^{T} \big(\log \pi_{\boldsymbol{\theta}}(y_t| \boldsymbol{x}, \boldsymbol{y}_{<t} ) - \log \pi^{*}(y_t| \boldsymbol{x}, \boldsymbol{y}_{<t} ) \big) \, \nabla_{\boldsymbol{\theta}} \log \pi_{\boldsymbol{\theta}}(y_t| \boldsymbol{x}, \boldsymbol{y}_{<t} ) 
\Big].
\end{aligned}
\end{equation}
Similarly, the approximated gradient of G-OPD in Eq.~(\ref{eq:generalized_opd}) can be written as
\begin{equation}
\begin{aligned}
\nabla_{\boldsymbol{\theta}} \mathcal{J}_{\text{G-OPD}}(\boldsymbol{\theta}) = 
\mathbb{E}_{\boldsymbol{x} \sim D, \boldsymbol{y} \sim \pi_{\boldsymbol{\theta}} (\cdot | \boldsymbol{x}) }\Big[
\sum_{t=1}^{T} A_{t}^{\text{G-OPD}} \, \nabla_{\boldsymbol{\theta}} \log \pi_{\boldsymbol{\theta}}(y_t| \boldsymbol{x}, \boldsymbol{y}_{<t} ) 
\Big],
\end{aligned}
\end{equation}
where $A_{t}^{\text{G-OPD}} = \big(\log \pi_{\boldsymbol{\theta}}(y_t| \boldsymbol{x}, \boldsymbol{y}_{<t} ) - \log \pi^{*}(y_t| \boldsymbol{x}, \boldsymbol{y}_{<t} ) \big) + (\lambda - 1)\big(\log \pi_{\text{ref}}(y_t| \boldsymbol{x}, \boldsymbol{y}_{<t} ) - \log \pi^{*}(y_t| \boldsymbol{x}, \boldsymbol{y}_{<t} ) \big)$.

\section{Detailed Training Settings}
\label{appendix: training settings}

The training hyper-parameters in math and code RL training are put in Table~\ref{tab: math RL hyper-parameters} and Table~\ref{tab: code RL hyper-parameters}  respectively. 

The training hyper-parameters in G-OPD in both domains are in Table~\ref{tab: G-OPD hyper-parameters}. In preliminary experiments, we find that under the same prompt size $*$ rollout $n$ conditions, setting a larger prompt size leads to smoother convergence. The number of optimization steps for G-OPD in all experiments with same-size teacher-student pairs (Section~\ref{subsec: same-sized distillation}) is set to 50, while it is set to 100 for experiments in the strong-to-weak distillation setting (Section~\ref{subsec: strong-to-weak distillation}). We find that further increasing the number of distillation steps may degrade generalization performance due to overfitting.

The training hyper-parameters in SFT are in Table~\ref{tab: SFT hyper-parameters}. We make sure the number of trajectories to each problem generated by the teacher in SFT is consistent with that generated by the student in OPD and ExOPD. We keep the number of optimization steps consistent with the corresponding G-OPD experiment for fair comparison.

\begin{table}[t]  
\begin{center}
\begin{minipage}[t]{0.48\linewidth}
\caption{Training hyper-parameters of GRPO in math RL.}
\label{tab: math RL hyper-parameters}
\centering
\begin{tabular}{ll}
\toprule
Hyper-parameter & Value \\
\midrule
Train Batch Size & 128  \\
Micro Batch Size & 128 \\
Rollout $n$ & 8 \\
Maximum Prompt Length & 2048 \\
Maximum Response Length & 16,384 \\
Temperature & 1.0 \\
Top-p & 1.0 \\
LR & $1\times 10^{-6}$ \\
Optimization Steps & 500 \\
KL Coefficient & 0.0 \\
\bottomrule
\end{tabular}
\end{minipage}  
\hfil
\begin{minipage}[t]{0.48\linewidth}
\caption{Training hyper-parameters of GRPO in code RL.}
\label{tab: code RL hyper-parameters}
\centering
\begin{tabular}{ll}
\toprule
Hyper-parameter & Value \\
\midrule
Train Batch Size & 128  \\
Micro Batch Size & 128 \\
Rollout $n$ & 8 \\
Maximum Prompt Length & 2048 \\
Maximum Response Length & 8192 \\
Temperature & 1.0 \\
Top-p & 1.0 \\
LR & $1\times 10^{-6}$ \\
Optimization Steps & 300 \\
KL Coefficient & 0.0 \\
\bottomrule
\end{tabular}
\end{minipage} 
\end{center}
\end{table}

\begin{table}[t]  
\begin{center}
\begin{minipage}[t]{0.48\linewidth}
\caption{Training hyper-parameters of G-OPD in both math and code domains.}
\label{tab: G-OPD hyper-parameters}
\centering
\begin{tabular}{ll}
\toprule
Hyper-parameter & Value \\
\midrule
Batch Size & 1024  \\
Rollout $n$ & 1 \\
Maximum Prompt Length & 2048 \\
Maximum Response Length & 16,384 \\
Temperature & 1.0 \\
Top-p & 1.0 \\
LR & $1\times 10^{-5}$ \\
\bottomrule
\end{tabular}
\end{minipage}  
\hfil
\begin{minipage}[t]{0.48\linewidth}
\caption{Training hyper-parameters of SFT in both math and code domains.}
\label{tab: SFT hyper-parameters}
\centering
\begin{tabular}{ll}
\toprule
Hyper-parameter & Value \\
\midrule
Batch Size & 1024  \\
Maximum Sequence Length & 32,768 \\
Warm-up Ratio & 0.05 \\
LR & $1\times 10^{-5}$ \\
\bottomrule
\end{tabular}
\end{minipage} 
\end{center}
\end{table}

\section{Results of Distillation from Domain Teachers with Sufficient RL Trainings}
\label{appendix: results under stronger teachers}
Here, we show the on-policy distillation results when the domain teachers are trained with sufficient RL steps (i.e., 1200 steps). The experimental settings are the same as that in the Section~\ref{subsec: same-sized distillation}. The results in Table~\ref{tab: results under stronger teachers} demonstrate the generalizability and effectiveness of ExOPD in this case.

\input{tables/results_under_stronger_teachers}

\section{Prompt Templates}
\label{appendix: prompt templates}
We show the prompt templates used in our experiments in the end.
\clearpage
\input{prompts/qwen3_math}
\input{prompts/qwen3_code}

\end{document}

%% file: math_commands.tex
\usepackage{amsmath,amsfonts,bm}
\usepackage{multicol}

\def\eqref#1{equation~\ref{#1}}

\def\1{\bm{1}}

\DeclareMathAlphabet{\mathsfit}{\encodingdefault}{\sfdefault}{m}{sl}
\SetMathAlphabet{\mathsfit}{bold}{\encodingdefault}{\sfdefault}{bx}{n}



%% file: tables/comparison_against_teacher_with_additional_rl.tex
\begin{table*}[t]
\caption{Comparison against the math domain teacher with continued RL. Each numerical subscript indicates the absolute \textcolor{lightgreen}{\textbf{improvement}} or \textcolor{myred}{\textbf{degradation}} \textbf{compared to the teacher model}.}
\label{tab: comparison with teacher with additional training}
\centering
\begin{tabular}{lccccc}
\toprule
\multirow{1}{*}{\begin{tabular}[c]{@{}l@{}}\textbf{Method} \end{tabular}}   & AIME24 & AIME25  & HMMT25 (Feb.) & HMMT25 (Nov.) & \phantom{0}\phantom{0}Avg.\phantom{0}\phantom{0}
\\
\midrule
\makecell[l]{Teacher}  &  58.0 & 54.6 & 32.5 & 38.9  & 46.0 \\
 ~~ \scriptsize{+ continued RL (100 steps)} & \cimp{60.9}{+2.9} & \cimp{55.6}{+0.5} & \cimp{32.8}{+0.3} &  \cdec{38.4}{-0.5} & \cimp{46.9}{+0.9} \\
\rowcolor{cyan!10} ExOPD (50 steps) & \cimp{\textbf{62.7}}{+4.7} & \cimp{\textbf{56.1}}{+1.5} & \cimp{\textbf{33.9}}{+1.4} & \cimp{\textbf{39.3}}{+0.4} & \cimp{\textbf{48.0}}{+2.0}  \\
\bottomrule
\end{tabular}
\end{table*}

%% file: tables/results_on_same_size_models.tex
\begin{table*}[t]
\caption{Comparison against off-policy distillation (SFT) and weight extrapolation (ExPO) methods in both single-teacher and multi-teacher settings with same-sized teacher-student pairs. ``Teacher'' represents the performance of the domain teacher model (Qwen3-4B-Non-Thinking-RL-Math for math reasoning and Qwen3-4B-Non-Thinking-RL-Code for code generation), ``Student'' represents the initial performance of student model Qwen3-4B-Non-Thinking. Each numerical subscript indicates the absolute \textcolor{lightgreen}{\textbf{improvement}} or \textcolor{myred}{\textbf{degradation}} \textbf{compared to the domain teacher model}.}
\label{tab: results on same-sized models}
\centering
\small
\setlength{\tabcolsep}{2.0pt}
\begin{tabular}{lccccccccc}
\toprule
\multirow{2.5}{*}{\begin{tabular}[c]{@{}l@{}}\textbf{Method} \end{tabular}} & \multicolumn{5}{c}{\textbf{Math Reasoning}} & \multicolumn{4}{c}{\textbf{Code Generation}}  \\
\cmidrule(lr){2-6}
\cmidrule(lr){7-10}
& AIME24 & AIME25  & HMMT25 (Feb.) & HMMT25 (Nov.) & \phantom{0}\phantom{0}Avg.\phantom{0}\phantom{0} & HumanEval+ & \phantom{0}MBPP+\phantom{0} & \phantom{0}\phantom{0}LCB\phantom{0}\phantom{0} & \phantom{0}\phantom{0}Avg.\phantom{0}\phantom{0}
\\
\midrule
\makecell[l]{Teacher}  &  58.0 & 54.6 & 32.5 & 38.9  & 46.0 & 86.0 & 70.2 & 27.3 & 61.2 \\
\makecell[l]{Student} & 21.5 & 21.9 & 10.0 & \phantom{0}8.0 & 15.4 & 74.7 & 64.7 & 17.9 & 52.4 \\
\midrule
\multicolumn{6}{l}{\emph{\quad \textbf{Single-Teacher Distillation}}} \\ 
ExPO & \cimp{58.7}{+0.7} & \cimp{55.2}{+0.6} & \cdec{32.4}{-0.1} & \cdec{37.0}{-1.9} & \cdec{45.8}{-0.2} & \cdec{84.8}{-1.2} & \cimp{70.2}{+0.0} & \cimp{28.0}{+0.7} & \cdec{61.0}{-0.2}\\
OPD &  \cimp{60.7}{+2.7} & \cimp{55.0}{+0.4} & \cdec{32.4}{-0.1} & \cdec{37.9}{-1.0} & \cimp{46.5}{+0.5} & \cdec{85.2}{-0.8} & \cdec{69.9}{-0.3} & \cimp{27.3}{+0.0} & \cdec{60.8}{-0.3} \\
\rowcolor{cyan!10} ExOPD & \cimp{\textbf{62.7}}{+4.7} & \cimp{\textbf{56.1}}{+1.5} & \cimp{\textbf{33.9}}{+1.4} & \cimp{\textbf{39.3}}{+0.4} & \cimp{\textbf{48.0}}{+2.0} & \cimp{\textbf{86.9}}{+0.9} & \cimp{\textbf{70.7}}{+0.5} & \cimp{\textbf{28.6}}{+1.3} & \cimp{\textbf{62.1}}{+0.9} \\
\midrule
\multicolumn{6}{l}{\emph{\quad \textbf{Multi-Teacher Distillation}}} \\ 
SFT & \cimp{58.5}{+0.5} &  \cdec{53.3}{-1.3} & \cdec{30.7}{-1.8} & \cdec{34.8}{-4.1} & \cdec{44.3}{-1.7} & \cimp{86.4}{+0.4} & \cdec{69.6}{-0.6} & \cdec{26.4}{-0.9} & \cdec{60.8}{-0.4} \\
ExPO & \cdec{57.5}{-0.5} & \cdec{54.5}{-0.1} & \cdec{31.7}{-0.8} & \cdec{36.3}{-2.6} & \cdec{45.0}{-1.0} & \cimp{\textbf{86.7}}{+0.7} & \cimp{\textbf{72.0}}{+1.8} & \cimp{\textbf{29.0}}{+1.7} &  \cimp{\textbf{62.6}}{+1.4} \\
OPD &  \cimp{60.6}{+2.6} &  \cdec{54.1}{-0.5} & \cimp{32.5}{+0.0} & \cdec{38.3}{-0.6} & \cimp{46.4}{+0.4} & \cdec{84.6}{-1.4} & \cdec{69.5}{-0.7} &  \cimp{27.6}{+0.3} & \cdec{60.6}{-0.6} \\
\rowcolor{cyan!10} ExOPD &  \cimp{\textbf{61.0}}{+3.0} &  \cimp{\textbf{56.0}}{+1.4} & \cimp{\textbf{34.4}}{+1.9}  &  \cimp{\textbf{39.2}}{+0.3} & \cimp{\textbf{47.7}}{+1.7}  & \cimp{86.3}{+0.3} & \cimp{70.6}{+0.4} &  \cimp{\textbf{29.0}}{+1.7} & \cimp{62.0}{+0.8} \\
\bottomrule
\end{tabular}
\end{table*}

%% file: tables/results_on_different_sized_models.tex
\begin{table*}[t!]
\caption{Evaluation accuracy on four math reasoning benchmarks in the strong-to-weak distillation setting. Teacher model is Qwen3-30B-A3B-Instruct-2507. The numerical subscript indicates the absolute \textcolor{lightgreen}{\textbf{improvement}} or \textcolor{myred}{\textbf{degradation}} \textbf{compared to the standard OPD}.}
\label{tab: math results on different-sized models}
\centering
\begin{tabular}{lccccc}
\toprule
  Method
& \phantom{0}AIME24\phantom{0} & \phantom{0}AIME25\phantom{0}  & HMMT25 (Feb.) & HMMT25 (Nov.) & \phantom{0}\phantom{0}Avg.\phantom{0}\phantom{0} \\
\midrule
Teacher  & 74.7 & 62.8 & 44.2 & 57.2 & 59.7 \\
\midrule
\multicolumn{6}{l}{\emph{\quad \textbf{Student: Qwen3-1.7B-Non-Thinking}}} \\ 
Base & 12.3 & 11.4 & \phantom{0}6.8 & \phantom{0}4.5 & \phantom{0}8.8 \\
SFT & 18.1 & 20.5 & \phantom{0}9.2 & \phantom{0}6.3 & 13.5 \\
OPD & 33.0 & 28.7 & 15.7 & 14.9  & 23.1  \\
\rowcolor{cyan!10} ExOPD & \cimp{\textbf{37.3}}{+4.3} & \cimp{\textbf{31.5}}{+2.8} & \cimp{\textbf{16.2}}{+0.5} & \cimp{\textbf{16.5}}{+1.6} & \cimp{\textbf{25.4}}{+2.3} \\
\midrule
\multicolumn{6}{l}{\emph{\quad \textbf{Student: Qwen3-4B-Non-Thinking}}} \\ 
Base &  21.5 & 21.9 & 10.0 & \phantom{0}8.0 & 15.4 \\
SFT & 45.4 & 40.9 & 22.4 & 31.6 & 35.1 \\
OPD & 55.0 & 48.0 & 29.8 & 37.7& 42.6  \\
\rowcolor{cyan!10} ExOPD & \cimp{\textbf{58.7}}{+3.7} & \cimp{\textbf{50.8}}{+2.8} & \cimp{\textbf{33.0}}{+3.2} & \cimp{\textbf{38.8}}{+1.1} & \cimp{\textbf{45.3}}{+2.7} \\ 
\bottomrule
\end{tabular}
\end{table*}

%% file: tables/results_under_stronger_teachers.tex
\begin{table*}[t]
\caption{Distillation results under stronger teachers with sufficient RL trainings. Each numerical subscript indicates the absolute \textcolor{lightgreen}{\textbf{improvement}} or \textcolor{myred}{\textbf{degradation}} \textbf{compared to the domain teacher model}.}
\label{tab: results under stronger teachers}
\centering
\small
\setlength{\tabcolsep}{2.0pt}
\begin{tabular}{lccccccccc}
\toprule
\multirow{2.5}{*}{\begin{tabular}[c]{@{}l@{}}\textbf{Method} \end{tabular}} & \multicolumn{5}{c}{\textbf{Math Reasoning}} & \multicolumn{4}{c}{\textbf{Code Generation}}  \\
\cmidrule(lr){2-6}
\cmidrule(lr){7-10}
& AIME24 & AIME25  & HMMT25 (Feb.) & HMMT25 (Nov.) & \phantom{0}\phantom{0}Avg.\phantom{0}\phantom{0} & HumanEval+ & \phantom{0}MBPP+\phantom{0} & \phantom{0}\phantom{0}LCB\phantom{0}\phantom{0} & \phantom{0}\phantom{0}Avg.\phantom{0}\phantom{0}
\\
\midrule
\makecell[l]{Teacher}  &  68.2 & 59.3 & 37.3 & 42.9 & 51.9 & 88.9 & 72.5 & 28.0 &  63.1 \\
\makecell[l]{Student} & 21.5 & 21.9 & 10.0 & \phantom{0}8.0 & 15.4 & 74.7 & 64.7 & 17.9 & 52.4 \\
\midrule
\multicolumn{6}{l}{\emph{\quad \textbf{Single-Teacher Distillation}}} \\ 
OPD & \cimp{68.3}{+0.1} & \cdec{58.7}{-0.6} & \cimp{\textbf{38.7}}{+1.4} & \cdec{41.2}{-1.7} & \cdec{51.7}{-0.2} & \cimp{89.3}{+0.4} & \cdec{71.3}{-1.2} & \cimp{28.0}{+0.0} &  \cdec{62.9}{-0.2} \\
\rowcolor{cyan!10} ExOPD & \cimp{\textbf{68.4}}{+0.2} & \cdec{\textbf{59.2}}{-0.1} & \cimp{38.2}{+0.9} & \cdec{\textbf{42.8}}{-0.1} & \cimp{\textbf{52.2}}{+0.3} & \cimp{\textbf{89.9}}{+1.0} & \cimp{\textbf{73.7}}{+1.2} & \cimp{\textbf{29.3}}{+1.3} &  \cimp{\textbf{64.3}}{+1.2}  \\
\midrule
\multicolumn{6}{l}{\emph{\quad \textbf{Multi-Teacher Distillation}}} \\ 
OPD & \cimp{68.2}{+0.0} & \cimp{\textbf{60.2}}{+0.9} & \cimp{\textbf{38.5}}{+1.2} & \cdec{40.8}{-2.1} & \cimp{51.9}{+0.0} & \cdec{86.4}{-2.5} & \cdec{72.1}{-0.4} &  \cdec{27.6}{-0.4} & \cdec{62.0}{-1.1}  \\
\rowcolor{cyan!10} ExOPD & \cimp{\textbf{70.1}}{+1.9} & \cimp{59.6}{+0.3} & \cimp{37.5}{+0.2} & \cdec{\textbf{42.7}}{-0.2} & \cimp{\textbf{52.5}}{+0.6} & \cimp{\textbf{89.5}}{+0.6} & \cimp{\textbf{73.9}}{+1.4} & \cimp{\textbf{29.7}}{+1.7} &  \cimp{\textbf{64.4}}{+1.3}  \\
\bottomrule
\end{tabular}
\end{table*}

%% file: prompts/qwen3_math.tex
\begin{prompt}[title={Training and Evaluation Prompt Template for Math Reasoning}]
\label{prompt: qwen3 math}
$<|$im\_start$|>$user
\\
\{question\}
\\
Please reason step by step, and put your final answer within  \verb|\|boxed\{\}.$<|$im\_end$|>$
\\
$<|$im\_start$|>$assistant
\end{prompt}

%% file: prompts/qwen3_code.tex
\begin{prompt}[title={Training and Evaluation Prompt Template for Code Generation}]
\label{prompt: qwen3 code}
$<|$im\_start$|>$user
\\
\{question\}
\\
Write Python code to solve the problem. Present the code in 
\\
\texttt{```}python
\\
Your code
\\
\texttt{```}
\\
at the end.
\\
You need to think first then write the Python code.$<|$im\_end$|>$
\\
$<|$im\_start$|>$assistant
\end{prompt}